%% file: main_unmarked.tex
\documentclass[a4paper,12pt]{article}

\usepackage[english]{babel}
\usepackage[utf8]{inputenc}
\usepackage{amsmath}
\usepackage{graphicx}
\usepackage{soul}
\usepackage[margin=1in]{geometry}
\usepackage[colorinlistoftodos]{todonotes}
\usepackage[bookmarksopen,bookmarksnumbered,colorlinks,linkcolor=blue,citecolor=blue]{hyperref}
\usepackage{amssymb}
\usepackage{gensymb}
\usepackage {tikz}
\usepackage{adjustbox}
\usepackage{multirow}
\usepackage{graphicx}
\usepackage[normalem]{ulem}
\usepackage{booktabs}      
\usepackage{multirow}      
\usepackage{threeparttable} 
\usepackage{bm} 
\usepackage{makecell}

\usepackage{apacite}
\usepackage{natbib}

\usepackage{xcolor}

\usetikzlibrary {positioning}

\title{A Bayesian latent class reinforcement learning framework to capture adaptive, feedback-driven travel behaviour}

\bigskip
\usepackage{float}
 \author{
    \textit{Georges Sfeir$^{1}$}, 
    \textit{Stephane Hess$^{1}$},
    \textit{Thomas O. Hancock$^{1}$},
    \textit{Filipe Rodrigues$^{2}$}, \\
    \textit{Jamal Amani Rad$^{1}$}, 
    \textit{Michiel Bliemer$^{3}$}, 
    \textit{Matthew Beck$^{3}$}, 
    \textit{Fayyaz Khan$^{4}$}
    \\ \\
    {\footnotesize
        \textit{$^1$Choice Modelling Centre \& Institute for Transport Studies, University of Leeds}}
        \\
    {\footnotesize
        \textit{$^2$DTU Management, Technical University of Denmark}}
        \\
    {\footnotesize
        \textit{$^3$Institute of Transport and Logistics Studies, University of Sydney}}
        \\
        {\footnotesize
        \textit{$^4$College of Business, Al Yamamah University, Riyadh, Saudi Arabia}}
    }

\date{}

\begin{document}
\maketitle
\section*{Abstract}
\input{Abstract_unmarked}
\newpage

\section{Introduction}
\input{Introduction_unmarked}

\section{Methodology}
\label{section:methodology}
\input{Methodology_unmarked}

\section{Dataset}
\label{section:dataset}
\input{Data_unmarked}

\section{Results}
\label{section:results}
\input{Results_unmarked}

\section{Conclusions}
\label{section:conclusion}
\input{Conclusions_unmarked}

\clearpage
\section*{Acknowledgment}
\input{Acknowledgment_unmarked}

\section*{Data availability}
The authors do not have permission to share data.

\clearpage
\appendix
\section*{Appendix A - Generative process of the standard RL model}
\label{section:appendixA}
\input{Appendix1_unmarked}

\clearpage
\section*{Appendix B - Synthetic validation and parameter recovery study}
\label{section:appendixB}
\input{Appendix2_unmarked}

\clearpage

\input{References_unmarked.bbl}
\end{document}

%% file: Abstract_unmarked.tex
Many travel decisions involve a degree of experience formation, where individuals learn their preferences over time. At the same time, there is extensive scope for heterogeneity across individual travellers, both in their underlying preferences and in how these evolve. The present paper puts forward a Latent Class Reinforcement Learning (LCRL) model that allows analysts to capture both of these phenomena. We apply the model to a driving simulator dataset and estimate the parameters through Variational Bayes. We identify three distinct classes of individuals that differ markedly in how they adapt their preferences: the first displays context-dependent preferences with context-specific exploitative tendencies; the second follows a persistent exploitative strategy regardless of context; and the third engages in an exploratory strategy combined with context-specific preferences.

\subsection*{Keywords}
Econometrics; Discrete choice models; Mathematical psychology; Reinforcement learning; Bayesian estimation

%% file: Introduction_unmarked.tex
Understanding how individuals make travel decisions in dynamic and uncertain contexts is a fundamental challenge in travel behaviour research. Traditional methods, such as Discrete Choice Models (DCMs), are mainly developed to capture preferences as fixed constructs \citep{Cirillo01072011}, even though travellers often revise their expectations as they accumulate new experiences. In many real-life scenarios, individuals might experience periods of adaptive learning, where they receive feedback (e.g., unexpected delays, newly available options) and adjust their expectations and preferences about the alternatives they face. Several factors can trigger such a feedback-based adaptive process. These include disruptions or temporary restrictions on preferred travel modes or routes, prompting travellers to explore alternative options. Additionally, the introduction of new transport alternatives — such as newly established transit routes, bike-sharing schemes, or emerging mobility services — can lead individuals to reconsider their habitual travel patterns and update their expectations, as they learn how these alternatives perform in practice. Life events and significant contextual changes, including relocation to a new residence, starting a new job, or moving to a different city or country, also frequently necessitate a period of exploration and adaptation as travellers familiarise themselves with the available transportation network and identify options that best fit their needs.

Such exploratory behaviour driven by these varied factors can initially result in instability and variability in travel choices observed in data. Yet, as travellers gain experience and confidence in their chosen options, their decisions might stabilise, resulting in preference stability driven by exploitation and, potentially, behavioural inertia  \citep{RAD2025}. While existing studies often emphasise the identification and quantification of stable behavioural patterns \citep{HESS2011973}, such as inertia and habitual behaviour \citep{GAO2020272, CANTILLO2007195, CHERCHI2011679}, they typically overlook the underlying mechanisms through which these stable preferences may have initially emerged. Thus, explicitly modelling and comprehensively understanding how travellers dynamically learn their preferences from experienced feedback remains relatively understudied \citep{RAD2025, HENRIQUEZJARA2025100552, HENRIQUEZJARA2025100877}. Several factors contribute to this research gap, including the lack of appropriate longitudinal datasets specifically designed to capture feedback-driven learning, practical difficulties in capturing dynamic learning behaviours using conventional travel surveys that typically record choices but not the experienced feedback/outcomes that shape them, and limitations of conventional travel modelling methods. This contrasts notably with the field of mathematical and behavioural psychology, where several models have been developed specifically to study the feedback-based learning and evolution of preferences over time \citep{NIV2009139, hBayesDM, GERSHMAN201834, FONTANESI201926, COLLINS2022104}. 

A well established theory in psychology and neuroscience is Reinforcement Learning (RL), which describes how individuals adjust their expectations and behaviour through experience-based feedback \citep{NIV2009139, SUTTON2018}. Within this framework, mathematical psychology provides formal RL models that operationalise this learning process to capture how individuals’ preferences and behaviours evolve over time based on observed/received feedback (positive or negative). In decision-making contexts, RL assumes that individuals form and update estimates of the value of each option/alternative based on experience. After making a choice, individuals adjust their expected value (e.g., expected travel time or travel cost) of the chosen alternative to reduce the prediction error based on the experienced outcome. This adjustment or learning process is driven by reward prediction errors (RPEs) \citep{COLLINS2022104, SAMSON201091}, defined as the difference between the reward an individual expected and the reward he/she actually obtained. Through repeated choices, and as such repeated feedback, this RPE-driven learning process allows individuals to reduce prediction errors, reinforcing behaviours/choices that yield better-than-expected outcomes and discouraging those that result in worse-than-expected ones \citep{LIU2015}.
In RL, exploitation refers to choosing the option with the currently highest learned (expected) value, whereas exploration refers to sometimes choosing options that are currently less valued or more uncertain, for instance to learn more about their outcomes and potentially improve future decisions. Such models have been widely implemented in several behavioural contexts such as gambling and risk-taking \citep{Steingroever2014161}, social learning \citep{Diaconescu2017618, Velez2021110}, addiction \citep{LEI20222124}, amongst others.

In the context of travel behaviour, RL can provide a framework to model how travellers dynamically update their preferences based on past travel times, unexpected delays, or new information. A recent paper by \citet{RAD2025} has demonstrated that Reinforcement Learning could accurately capture behaviour in two travel behaviour datasets, uncovering the underlying learning process of travellers’ preferences. Their specification incorporates heterogeneity through a hierarchical Bayesian structure, where individual-level cognitive parameters are drawn from single-component Normal or Lognormal distributions. This approach is well suited to capturing continuous variation around a common behavioural process. However, when the objective is to uncover qualitatively distinct learning types that might exist in the population, such as persistent explorers who repeatedly try unfamiliar options versus rapid exploiters who quickly converge to more habitual behaviour, a discrete representation of heterogeneity can provide a complementary perspective. In particular, a unimodal hierarchical prior is not designed to represent multimodal behavioural patterns without introducing additional mixture components. As a result, modelling all individuals as varying continuously around a single population-level process may obscure behaviourally meaningful segments and limit the targeting of policy interventions, for example when information strategies benefit explorers but hinder exploiters. To address this, we introduce a hybrid Latent Class Reinforcement Learning (LCRL) framework for modelling adaptive, feedback-driven travel behaviour under discrete behavioural heterogeneity. The central contribution of the paper is the integration of latent class choice models (LCCMs) \citep[see e.g.,][]{Hess2024LCCM} with class-specific reinforcement-learning dynamics, allowing qualitatively distinct traveller types to differ not only in their route preferences, but also in how they update expectations from feedback and how they balance exploration and exploitation over time. In this framework, the latent class structure represents discrete forms of unobserved heterogeneity in learning behaviour, while the reinforcement learning component captures the evolution of expectations within each class through experience-based prediction errors.
A second contribution is computational. The proposed model is estimated in a fully Bayesian framework using variational inference \citep{Jordan1999183, Wainwright20081}, which provides an approximate Bayesian estimation strategy, yielding variational posterior distributions over all model parameters and associated uncertainty summaries, while enabling a scalable estimation \citep{RODRIGUES20221} despite the high dimensionality introduced by dynamic learning and class segmentation. Empirically, we apply the model to a unique dataset that combines repeated route choices in a driving simulator with responses from a stated-preference survey. This design allows travellers to receive and react to both experienced and hypothetical route choice feedback, offering an opportunity to track how expectations evolve across contexts that vary in realism and cognitive effort. The proposed LCRL model, integrating reinforcement learning principles from mathematical psychology with latent class choice modelling from econometrics, is applied to this data to jointly capture experience-based learning and discrete forms of behavioural heterogeneity. In doing so, it provides a scalable and richer characterisation of how different types of travellers learn from feedback, explore alternatives, and potentially converge towards more stable expectations and choice patterns over time. More broadly, the framework is relevant for emerging transport contexts in which travellers learn from repeated interaction with changing systems, such as real-time information provision, new mobility services, automated vehicles, disruptions, or newly introduced infrastructure.

The remainder of the paper is organised as follows. First, we introduce the theory behind RL and present the modelling and estimation methodology of the proposed LCRL model in Section \ref{section:methodology}. Next, we discuss the driving simulator route choice case study in Section \ref{section:dataset} and present the results in Section \ref{section:results}. Finally, we conclude in Section \ref{section:conclusion}.

%% file: Methodology_unmarked.tex
This section presents an overview of Reinforcement Learning from mathematical psychology (Section \ref{subsection2:RL}), an illustrative example (Section \ref{subsection2:RLExample}), and the formulation of the proposed Latent Class Reinforcement Learning (LCRL) model (Section \ref{subsection2:LCRL}). The LCRL model is presented in a general form,   enabling its application within travel behaviour and potentially in other domains as well (e.g., healthcare, marketing and consumer behaviour, economics, etc.).

\subsection{Reinforcement learning}
\label{subsection2:RL}
Reinforcement Learning, from mathematical psychology, models how individuals learn from feedback over time. The assumption is that when facing a choice situation, individuals have prior expectations of the available/considered alternatives \citep{SUTTON2018}. After choosing an alternative, they receive some sort of feedback (outcome), update their expectation about the chosen alternative based on the difference between their previous expectation and the received feedback (outcome), and then use the updated expectation for the next choice situation. This update of expectations, based on experience, is known as Reinforcement Learning (RL). There are several RL models, each offering different perspectives on how individuals learn from experience. This paper relies on the Rescorla-Wagner model \citep{Rescorla1972}, a framework that explains how values of alternatives evolve over time based on the difference between expected and actual outcomes, also known as prediction errors \citep{Schultz1997, ODoherty2003}. The Rescorla-Wagner model consists of two sub-components, the prediction error and the choice rule.

\subsubsection{The prediction error}
The prediction error $\delta$, defined as the difference between the expected outcome $Q$ and the actual outcome $r$, drives updates to future expectations $Q$. As such, the expectation of an individual $n\in\{1,\dots,N\}$ for an alternative $i\in\{1,\dots,I\}$ at trial $t\in\{1,\dots,T_{n}\}$ would evolve trial-by-trial as follows:
\begin{equation}
\label{eq:update_Q}
    Q_{nit} = Q_{ni(t-1)} + \alpha\delta_{ni(t-1)} = Q_{ni(t-1)} + \alpha(r_{ni(t-1)} - Q_{ni(t-1)}),
\end{equation}
where $Q_{ni(t-1)}$ is the prior expectation, $\alpha$ is the learning rate, $\delta_{ni(t-1)}$ is the prediction error of alternative $i$ made by individual $n$ at time $t-1$, and $r_{ni(t-1)}$ is the actual outcome (feedback) of alternative $i$ at time $t-1$. Note that, in-line with the standard Rescorla–Wagner chosen-only formulation, only the expectation of the alternative selected at each trial is updated. Alternative reinforcement learning models allow for fictive updates to unchosen alternatives or incorporate decay of unchosen values, but these extensions are not considered in the present formulation. In addition, $Q_{ni0}$ represents the initial expectation assigned by individual $n$ to alternative $i$ at $t=0$ before any outcomes are observed. These initial expectations can be interpreted as prior beliefs about the value of each alternative, which are subsequently updated through experience-driven feedback (Eq. \ref{eq:update_Q}). 

The learning rate $\alpha$ ranges between 0 and 1, where higher values mean that individuals place greater weight on recent feedback when updating their beliefs or $Q$-values. This means that individuals would quickly incorporate new information into their estimates and heavily discount older experiences. Correspondingly, lower learning rates mean that less weight is placed on recent outcomes and expectations are updated slowly by retaining older experiences.

\subsubsection{The choice rule}
\label{subsubsection:choice_rule}
The prediction errors are then transformed to choice probabilities by embedding them within a Random Utility Maximisation (RUM) style framework. Specifically, the utility of an individual $n$ choosing alternative $i$ at time $t$ is assumed to depend on the individual’s corresponding expectation $Q_{nit}$. As such, the utility is specified as:
\begin{equation}
    U_{nit} = \gamma_{i} \pm \beta Q_{nit} + \varepsilon_{nit},
\end{equation}
where $\gamma_{i}$ is the alternative-specific constant (or bias parameter) of alternative $i$, and $\beta$ is a non-negative sensitivity parameter that controls the exploration-exploitation trade-off behaviour of individuals. Higher $\beta$ values encourage exploitation behaviour while lower values lead to more exploration. The sign ($\pm$) depends on whether the expected outcome $Q_{nit}$ represents a reward (positive sign) or a cost (negative sign).

At this stage, it is assumed that individuals share the same learning rate and sensitivity parameter. Moreover, it is assumed that all individuals share a common initial expectation for each alternative, denoted $Q_{i,0}$ such that $Q_{ni0} = Q_{i,0}$ for all individuals $n$. These simplifying assumptions allow us to focus on the general learning dynamics captured by the model without introducing individual-level heterogeneity at this stage. Heterogeneity is later introduced in the LCRL extension via class-specific parameters.

Assuming the random error term $\varepsilon_{nit}$ is independently and identically distributed (iid) Extreme Value Type I over individuals and alternatives, the probability of individual $n$ choosing alternative $i$ at trial $t$ follows the conventional Multinomial Logit (MNL) formulation and is defined as:

\begin{equation}
\label{eq:mnl_p}
P_{nit} \;=\; P\bigl(y_{nt} = i \,\bigm|\,
    \bm{Q_{0}}, \bm{\gamma},\, \beta,\, \alpha\bigr)
\;=\;
\frac{\exp\bigl(\gamma_i \,\pm\, \beta\,Q_{nit}\bigr)}%
     {\displaystyle \sum_{i'=1}^{I}
        \exp\Bigl(\gamma_{i'} \,\pm\, \beta\,Q_{ni't}\Bigr)},
\end{equation}
where $\bm{Q_{0}} = \{Q_{1,0},\ldots,Q_{I,0}\}$, $\bm{\gamma} = \{\gamma_1,\ldots,\gamma_{I}\} $, with one $\gamma_i$ set to zero for identification, and
$y_{nt} \in \{1,\ldots,I\}$ captures the observed choice by individual $n$ in situation $t$.
The probability of observing all choices of an individual $n$, $\bm{y}_{n}$, during all time periods (scenarios) $T_{n}$ is then expressed as follows:
\begin{equation}
P\bigl(\bm{y}_{n} \mid \bm{Q_{0}}, \bm{\gamma}, \beta,\alpha\bigr)
= \prod_{t=1}^{T_n}
      P_{ny_{nt}t}
\end{equation}

\subsection{Reinforcement learning example}
\label{subsection2:RLExample}
To better illustrate the mechanics of the Rescorla-Wagner Reinforcement Learning framework and its connection to choice modelling, consider an individual $n$ who must choose every day between two unfamiliar routes, A and B. On the first day/trial ($t=0$), individual $n$ holds equal prior expectations for both alternatives, assigning the same expected travel time to each ($Q_{nA0} = Q_{nB0} = 25$). However, a baseline preference is assumed for route B ($\gamma_B = 1$) over route A ($\gamma_A = 0$). The sensitivity to the expected travel time is fixed to $\beta = 1$, balancing exploration and exploitation across all trials. It also enters the utility function negatively as the expectations in this case represent travel time (a cost), reflecting a preference for shorter travel times. Under these assumptions, the utilities of individual $n$ choosing alternatives A and B, respectively, at trial 0 are: 
\begin{itemize}
  \item $U_{nA0} = \gamma_A - \beta Q_{nA0} = 0 -1\times25 = -25$,
  \item $U_{nB0} = \gamma_B - \beta Q_{nB0} = 1 -1\times25 = -24$.
\end{itemize}
Using Eq. \ref{eq:mnl_p}, the probabilities of individual $n$ choosing alternatives A and B at trial 0, $P_{nA0}$ and $P_{nB0}$, would be 0.27 and 0.73, respectively. Suppose that individual $n$ chooses alternative B, but due to unexpected delays, experiences a travel time of 30 minutes ($r_{nB0}$). This leads to a prediction error of $\delta_{nB0} = r_{nB0} - Q_{nB0} = 5$ minutes. The extent to which this experience affects future expectations of alternative B depends on the learning rate (Eq.~\ref{eq:update_Q}).
Table~\ref{T1} demonstrates how expectations and choice probabilities would evolve under two different learning rate scenarios at trial $t = 1$: high learning rate ($\alpha$ = 0.9) and low learning rate ($\alpha$ = 0.1). 
\begin{table}[h]
    \caption{Expectations and probabilities at $t=1$}
    \label{T1}
    \centering
    \begin{tabular}{l l l}
        \hline
        $t = 1$ & High Learning Rate ($\alpha$ = 0.9) & Low Learning Rate ($\alpha$ = 0.1)\\ \hline
        $Q_{nA1}$ & 25 (not updated) & 25 (not updated) \\ 
        $Q_{nB1}$ & $25+0.9\times5$ = 29.5 & $25+0.1\times5$ = 25.5 \\ 
        $U_{nA1}$ & $-25$ & $-25$ \\ 
        $U_{nB1}$ & $-28.5$ & $-24.5$ \\ 
        $P_{nA1}$ & 0.97 & 0.38 \\ 
        $P_{nB1}$ & 0.03 & 0.62 \\ \hline
    \end{tabular}
\end{table} \\

If individual $n$ has a high learning rate ($\alpha = 0.9$), they would quickly revise their expectation of the previously chosen alternative, leading to sharp decline in the utility of route B ($U_{nB1}$ = -28.5) and its choice probability ($P_{nB1}$ = 0.03). In such a scenario, individual $n$ would be overwhelmingly inclined to switch to route A at $t=1$. In contrast, under a low learning rate ($\alpha = 0.1$), the expectation of route B would remain close to its original value ($Q_{nB1}$ = 25.5), leading to a modest change in its utility ($U_{nB1}$ = -24.5) and choice probability ($P_{nB1}$ = 0.62), and keeping route B as the most attractive alternative. In this case, individual $n$ would still be inclined to choose the initially preferred route B while gradually accumulating new information. This example highlights how reinforcement learning models capture individuals' response to feedback. A high learning rate leads to rapid adaptation based on new experiences, promoting exploratory behaviour and overreaction to recent experiences. In contrast, a low learning rate leads to slower but gradual adaptation to recent feedback, reinforcing habitual or previously preferred choices, and reducing overreaction to recent negative feedback.

\subsection{Latent class reinforcement learning (LCRL) model}
\label{subsection2:LCRL}
The proposed model combines Latent Class (LC) models from econometrics with class-specific Reinforcement Learning (RL) from mathematical psychology to capture both distinct behavioural patterns via latent classes and dynamic feedback for decision-making using RL within each class. While latent class models are widely used in econometrics to represent discrete forms of heterogeneity, and RL models have been applied in mathematical psychology to describe experience-based learning, their integration into a single framework remains rare. To the best of our knowledge, this is the first application that jointly models discrete behavioural heterogeneity and dynamic reinforcement learning using a fully Bayesian variational approach. This hybrid Latent Class Reinforcement Learning (LCRL) model thus allows to identify qualitatively different learning types, such as exploratory versus exploitative decision-makers, while accounting for the temporal evolution of preferences within each latent class.
\subsubsection{Latent class membership}
The latent class membership probabilities capture the likelihood that individual $n$ belongs to a class $k\in\{1,\dots,K\}$ as a function of socio-demographic and socio-economic characteristics of individuals and context attributes. These probabilities are obtained from an underlying membership function defined as:
\begin{equation}
    I_{nk} = \bm{X}^{\top}_{n}\bm{\eta}_{k} + \upsilon_{nk},
\end{equation}
where $\bm{X}_{n}$ is a vector of characteristics of individual $n$ and context attributes including a constant, $\bm{\eta}_{k}$ is a vector of corresponding unknown parameters, and $\upsilon_{nk}$ is a random disturbance term. For identification, the parameters for one class are fixed to zero, say $\bm{\eta}_K=0$.

Under the assumption that the disturbance terms are iid Extreme Value Type I over individuals and classes, the probability of an individual $n$ belonging to class $k$ is then defined as:
\begin{equation}
\label{eq:P_class}
P\bigl(s_{n} = k | \bm{X}_{n}, \bm{\eta}\bigr)
= \frac{\exp\bigl(\bm{X}^{\top}_{n}\bm{\eta}_{k}\bigr)}{\displaystyle \sum_{j=1}^K \exp\bigl(\bm{X}^{\top}_{n}\bm{\eta}_{j}\bigr)},
\end{equation}
where $s_{n} \in \{1,\ldots,K\}$ and $\bm{\eta} =(\bm{\eta}_{1},\ldots,\bm{\eta}_{K})$.

\subsubsection{Class-specific RL choice model}
The class-specific choice component of the proposed model follows the reinforcement learning structure presented in Section \ref{subsection2:RL} by assuming that the two sub-components of the Rescorla-Wagner model, the prediction error and the choice rule, are class specific. For completeness, we restate the class-specific version of the RL choice model below. Although this closely parallels the standard RL formulation in Section \ref{subsection2:RL}, presenting the equations with explicit class indices helps to introduce the notation used later in the generative process and ensures clarity for readers less familiar with reinforcement learning models.
\\ \\
\uline{The prediction error}
\\ \\
Within each class, an individual’s expectation of an alternative $i$ at trial $t$, conditioned on belonging to class $k$, would evolve trial-by-trial as follows:
\begin{equation}
Q_{nitk}=Q_{ni(t-1)k} + \alpha_{k}\delta_{ni(t-1)k} = Q_{ni(t-1)k} + \alpha_{k}\Bigl(r_{ni(t-1)} - Q_{ni(t-1)k}\Bigr),
\end{equation}
where $\alpha_{k}$ is the learning rate of individuals belonging to class $k$, $\delta_{ni(t-1)k}$ the prediction error of alternative $i$ made by individual $n$ at time $t-1$, and $r_{ni(t-1)}$ the actual outcome (feedback) of alternative $i$ at time $t-1$ experienced by individual $n$. Note that the initial expectations $Q_{i0k}$ are treated as free parameters and estimated from the observed data, rather than being imposed a priori.
\\ \\
\uline{The choice rule}
\\ \\
Within each latent class, the prediction error learning process is transformed to choices through the RUM-style framework introduced in Section \ref{subsubsection:choice_rule}. The expected values $Q_{nitk}$ are treated as class-specific systematic components of utility, combined with two class-specific cognitive parameters from mathematical psychology, bias $\gamma_{ik}$ and exploration-exploitation $\beta_{k}$ parameters. Conditioned on class $k$, the probability of individual $n$ choosing alternative $i$ at trial $t$ is:
\begin{equation}
P_{nitk}
= P\bigl(y_{nt} = i | \bm{Q}_{0k}, \bm{\gamma}_{k},
   \beta_{k}, \alpha_{k}, s_{n} = k \bigr)
= \frac{\exp\bigl(\gamma_{ik}\pm\, \beta_{k}\,Q_{nitk}\bigr)}
       {\displaystyle \sum_{i'=1}^{I}\exp\Bigl(\gamma_{i'k}\pm\, \beta_{k}\,Q_{ni'tk}\Bigr)},
\end{equation}
where $\bm{Q}_{0k} = \{Q_{10k},\ldots,Q_{I0k}\}$, $\bm{\gamma}_{k} = \{\gamma_{1k},\ldots,\gamma_{Ik}\} $, with  one $\gamma_{ik}$ per class $k$ normalised to zero for one alternative $i$ for identification, and $y_{nt} \in \{1,\ldots,I\}$ captures the observed choices.

Conditional on class $k$, the probability of observing the full sequence of choices $\bm{y}_n = (y_{n1},\ldots,y_{nT_n})$ made by individual $n$ over the $T_n$ trials is:
\begin{equation}
\label{eq:P_choice_class}
P\bigl(\bm{y}_n | \bm{Q}_{0k}, \bm{\gamma}_{k},
   \beta_{k}, \alpha_{k}, s_{n} = k \bigr)
= \prod_{t=1}^{T_n} 
    P_{n y_{nt} t k}.
\end{equation}

\subsubsection{Bayesian LCRL: joint model and likelihood}
Let $\bm{\Theta} = \{\bm{s},\bm{\eta},\bm{\beta},\bm{\gamma},\bm{\alpha},\bm{Q_{0}}\}$ denote the set of all latent variables across alternatives and/or classes in the proposed LCRL model that need to be estimated. The unconditional probability (or likelihood) of the observed choices of individual $n$ can be obtained by mixing the conditional reinforcement learning choice probability (Eq.~\ref{eq:P_choice_class}) over the probability of belonging to each class $k$ (Eq.~\ref{eq:P_class}). Then, assuming the availability of a sample of independent individuals, the likelihood over all $N$ individuals is:
\begin{equation}
P(\bm{Y} | \bm{\Theta})
= \prod_{n=1}^N 
  \sum_{k=1}^K 
    P\bigl(s_{n} = k | \bm{X}_{n}, \bm{\eta}\bigr)\;
    P\bigl(\bm{y}_n | \bm{Q}_{0k}, \bm{\gamma}_{k},
   \beta_{k}, \alpha_{k}, s_{n} = k \bigr),
\end{equation}
where $\bm{Y} = \{\bm{y}_1,\ldots,\bm{y}_N\}$ is the set of observed choices for all individuals.

All unknown parameters and latent variables are specified in a fully-Bayesian manner. The generative process of the fully Bayesian LCRL model can then be summarised as follows:

\begin{enumerate}
  \item \textbf{For each latent class} $k \in \{1,\ldots,K - 1\}$:
    \begin{enumerate}
      \item Draw class assignment parameters $\bm{\eta}_{k} \;\sim\; \mathcal{N}\bigl(\mu_{\bm{\eta}_{k}},\, \Sigma_{\bm{\eta}_{k}}\bigr)$
    \end{enumerate}

  \item \textbf{For each latent class} $k \in \{1,\ldots,K\}$:
    \begin{enumerate}
      \item Draw class-specific RL exploration-exploitation parameter
            $\beta_{k} \;\sim\; \mathrm{Lognormal}\bigl(\mu_{\beta_{k}},\, \sigma_{\beta_{k}}\bigr)$,
ensuring that $\beta_{k} > 0$.
      \item Draw class-specific RL learning-rate parameter 
            $z_{\alpha_{k}} \;\sim\; \mathcal{N}\bigl(\mu_{z_{\alpha_{k}}},\, \sigma_{z_{\alpha_{k}}}\bigr)$,
            \\
            then apply a sigmoid transformation to ensure a rate in \([0,1]\):
            \\
            $\alpha_{k} \;=\; S\bigl(z_{\alpha_{k}}\bigr) 
                           \;=\; \frac{1}{1 \;+\; e^{-\,z_{\alpha_{k}}}}$
      \item \textbf{For each alternative} $i \in \{1,\ldots,I\}$:
        \begin{enumerate}
          \item Draw class-specific RL bias parameters
                $\gamma_{ik} \;\sim\; \mathcal{N}\bigl(\mu_{\gamma_{ik}},\, \sigma_{\gamma_{ik}}\bigr)$
                \\
                (One alternative per class is set to zero for identification purposes.)
          \item Draw initial expectations
                $z_{Q_{i0k}} \;\sim\; \mathcal{N}\bigl(\mu_{z_{Q_{i0k}}},\, \sigma_{z_{Q_{i0k}}}\bigr)$,
                \\
                then apply the following sigmoid transformation to ensure positive initial $Q$ values in \([a,b]\):
                $Q_{i0k} = a+(b-a) \times S\bigl(z_{Q_{i0k}}\bigr)$, where $a$ and $b$ are fixed and determined by the experimental design and are not estimated.
        \end{enumerate}
    \end{enumerate}

  \item \textbf{For each individual} $n \in \{1,\ldots,N\}$:
    \begin{enumerate}
      \item Draw the latent class assignment
            $s_{n} \;\sim\; \mathrm{MNL}\bigl(\bm{X_{n}},\,\{\bm{\eta}_{k}\}_{k=1}^K\bigr)$
          \item \textbf{For each choice occasion} $t \in \{1,\ldots,T_{n}\}$ and conditioned on the realised assignment $s_{n}=k$:
            \begin{enumerate}
                  \item Draw observed choice
                        $y_{nt} \;\sim\; \mathrm{MNL}\bigl(\beta_{k},\,\{\gamma_{ik},\,Q_{nitk}\}_{i=1}^I\bigr)$,
              \item Compute the prediction error of the chosen alternative $y_{nt} = j$ (according to step 3.(b).i):\vspace{-0.8em}
                    \[\delta_{njtk} 
                      \;=\; r_{njt} \;-\; Q_{njtk}\]
              \item Update the expectation of the chosen alternative $y_{nt} = j$ (according to step 3.(b).i):\vspace{-0.8em}
                    \[Q_{nj(t+1)k} \;=\; 
                    Q_{njtk}
                    \;+\; \alpha_{k}\,\delta_{njtk}\]
            \end{enumerate}
    \end{enumerate}
\end{enumerate}

\noindent Given the Bayesian generative process, the joint distribution of the LCRL model is defined as: 
\begin{align}
\label{eq:joint_p}
P(\bm{Y}, \bm{\Theta}) &= \prod_{k=1}^{K-1} 
P(\bm{\eta}_k) \left[ \prod_{k=1}^{K} P(\beta_{k}) P(\alpha_{k}) \left(\prod_{i=1}^{I-1} P(\gamma_{ik})\right) \prod_{i=1}^{I} P(Q_{i0k})\right]\notag \\
&\quad \times \prod_{n=1}^{N} \prod_{k=1}^{K} \Big[ P(s_{n} = k | \bm{X}_n, \bm{\eta})
P(\bm{y}_n | \bm{Q}_{0k}, \bm{\gamma}_{k}, \beta_{k}, \alpha_{k}, s_{n} = k) \Big]^{\mathbf{1}(s_{n} = k)},
\end{align}
where $\mathbf{1}(s_n=k)$ is an indicator equal to one if individual $n$ belongs to class $k$, and zero otherwise. Therefore, only the likelihood contribution corresponding to the realised class assignment enters the joint distribution for each individual.
Our goal is then to infer $\bm{\Theta}$ given the observed data. According to Bayes’ rule, the posterior distribution of $\bm{\Theta}$ is given by:
\begin{equation}
P(\bm{\Theta} | \bm{Y}) = \frac{P(\bm{Y} |\bm{\Theta})P(\bm{\Theta})}{\int P(\bm{Y} |\bm{\Theta})P(\bm{\Theta})\, d\bm{\Theta}},
\end{equation}
where the dependence on $\bm{\Theta}$ is explicit.
However, this posterior is computationally intractable due to the high-dimensional integral of the denominator (i.e., marginal likelihood). Hence, approximate inference methods are needed.

\subsubsection{Variational inference}
The goal of inference in the LCRL model is to estimate the latent variables and unknown parameters that drive individual decision-making. These include the latent class assignments $s_{n}$, the class membership parameters $\eta_{k}$ that link observed individual attributes to latent class membership, the class-specific reinforcement learning parameters—namely, the bias parameters $\gamma_{ik}$ and the exploration–exploitation parameters $\beta_{k}$—as well as the learning rate parameters which are parametrised via latent variables $z_{\alpha_{k}}$, and the initial $Q$-value parameters $z_{Q_{i0k}}$. 

Markov-chain Monte Carlo (MCMC) methods remain the standard Bayesian estimation approach for mixture models in both the choice modelling \citep{Train_2009, DANAF201935, RODRIGUES20221} and reinforcement learning \citep{RAD2025, hBayesDM} literatures. Their main advantage is that, under appropriate regularity and convergence conditions, they provide asymptotically exact posterior simulation. However, this comes at a substantial computational cost, particularly for high-dimensional posterior distributions, and requires careful assessment of convergence \citep{BLEI03042017, TAN201727, BANSAL2020124, RODRIGUES20221}. Variational Inference (VI) or Variational Bayes (VB) \citep{Jordan1999183, Wainwright20081} provides an alternative approximate Bayesian estimation strategy by transforming posterior inference into an optimisation problem. Rather than generating samples from the exact posterior, VI selects a tractable family of distributions and optimises its parameters so that the resulting variational distribution is close to the true posterior. This optimisation-based formulation can lead to substantial gains in computational efficiency. In the context of mixture choice models, previous studies have shown that VI can achieve speedups of several orders of magnitude relative to MCMC or traditional maximum simulated likelihood estimation  approaches, while delivering similar parameter recovery, predictive performance, and uncertainty summaries in the applications considered \citep{TAN201727, BANSAL2020124, RODRIGUES20221}. The trade-off is that the resulting posterior is approximate and may be affected by the choice of variational family.

This trade-off is particularly relevant for the proposed LCRL model, which combines latent class membership, class-specific reinforcement-learning parameters, and trial-by-trial Q-value updates. This structure creates a high-dimensional posterior and would make repeated estimation across alternative class specifications computationally demanding under MCMC. We therefore approximate the true posterior distribution $P(\bm{\Theta}|\bm{Y})$ using VI, by selecting a family of tractable distributions  $q_\phi (\bm{\Theta})$ parametrised by $\phi$. The inference problem is then transformed to an optimisation problem where the objective is to determine the parameters $\phi$ that allow the variational approximation $q_\phi (\bm{\Theta})$ to closely match the true posterior distribution $P(\bm{\Theta}|\bm{Y})$. In this framework, we introduce independent variational distributions over the latent class assignments $\bm{s}$ and latent variables $\{\bm{\eta},\bm{\beta},\bm{\gamma},\bm{\alpha},\bm{Q_{0}}\}$. To make the inference tractable, we adopt a mean-field variational approximation and assume that the variational posterior factorises across all latent variables:
\begin{equation}
    q_{\phi}(\bm{\Theta}) = \prod_{m}q_{\phi}(\Theta_{m}),
\end{equation} implying conditional independence among all components of $\bm{\Theta}$ in the variational approximation.  The variational distribution of the class assignments $q_\phi(s_{n})$ is modelled using a categorical distribution with parameters $\pi_{n}$, representing the probability that individual $n$ belongs to latent class $k$. The variational distribution for the class membership parameters $q_\phi(\eta_{k})$ is modelled using a multivariate normal distribution with a mean vector $\mu_{\eta_{k}}$ and a covariance $\Sigma_{\eta_k}$. Similarly, the variational distribution for the class-specific RL bias parameters $q_\phi(\gamma_{ik})$ are modelled as normal distributions parametrised by a mean vector $\mu_{\gamma_{ik}}$ and a standard deviation $\sigma_{\gamma_{ik}}$. The variational distribution for the class-specific RL exploration-exploitation parameters $q_\phi({\beta_{k}})$ is modelled as a lognormal distribution with parameters $\mu_{\beta_{k}}$ and $\sigma_{\beta_{k}}$, ensuring that $\beta_{k}$ is positive.
Finally, the latent variables driving the learning rate dynamics $z_{\alpha_{k}}$ and the initial $Q$-value $z_{Q_{i0k}}$ are modelled as normal distributions $q_\phi(z_{\alpha_{k}})$ and $q_\phi(z_{Q_{i0k}})$, respectively. These latent variables are subsequently transformed via sigmoid functions to ensure that the derived learning rates and initial $Q$-values fall within valid ranges.

The distance between the true and approximate posterior distributions can be measured by the Kullback–Leibler (KL) divergence:
\begin{equation}
\label{eq:kl_q_to_p}
\mathrm{KL}\big(q_{\phi}(\bm{\Theta}) \,\|\, P(\bm{\Theta} \mid \bm{Y})\big) = 
\int q_{\phi}(\bm{\Theta}) \log\left( \frac{q_{\phi}(\bm{\Theta})}{P(\bm{\Theta} | \bm{Y})} \right) \, d\bm{\Theta}
\end{equation}

However, the KL divergence above is intractable and therefore cannot be directly minimised. Instead, following the theory of VI \citep{Jordan1999183, Wainwright20081}, the variational parameters can be equivalently optimised by maximising the Evidence Lower Bound (ELBO), defined as:
\begin{equation}
\label{eq:elbo}
\mathrm{ELBO} = \mathbb{E}_{q_\phi} \left[ \log P(Y | \bm{\Theta}) \right] 
- \mathrm{KL}\big(q_\phi(\bm{\Theta}) \,\|\, P(\bm{\Theta})\big)
\end{equation}
where $\mathbb{E}_{q} \left[ \log P(Y \mid \bm{\Theta}) \right]$ is the expected log-likelihood of the model, and $\mathrm{KL}\big(q_\phi(\bm{\Theta}) \,\|\, P(\bm{\Theta})\big)$ is the Kullback–Leibler divergence that regularises the variational distributions of all unknown parameters $\bm{\Theta}$ to be close to their prior distributions. Using the joint distribution of the model (Eq. \ref{eq:joint_p}) together with the fully independent (mean-field) approximation assumed for $q_\phi(\bm{\Theta})$, the $\mathrm{KL}$ divergence term decomposes as follows:
\begin{align}
\mathrm{KL}(q_\phi(\bm{\Theta}) \,\|\, P(\bm{\Theta})) &= 
\mathrm{KL}(q_\phi(s_{n}) \,\|\, P(s_{n})) + 
\mathrm{KL}(q_\phi(\eta_k) \,\|\, P(\eta_k)) + 
\mathrm{KL}(q_\phi(\beta_k) \,\|\, P(\beta_k)) \notag \\
&\quad + \mathrm{KL}(q_\phi(\gamma_{ik}) \,\|\, P(\gamma_{ik})) + 
\mathrm{KL}(q_\phi(z_{\alpha_{k}}) \,\|\, P(z_{\alpha_{k}})) \notag \\
&\quad + \mathrm{KL}(q_\phi(z_{Q_{i0k}}) \,\|\, P(z_{Q_{i0k}}))
\end{align}

Maximising the ELBO with respect to the variational parameters $\phi$ is equivalent to minimising the KL divergence in Eq.~\ref{eq:kl_q_to_p}. We perform the optimisation of the ELBO using the Adam optimiser \citep{kingma2017adam}, a stochastic gradient-based method, allowing for efficient estimation of the variational parameters in high-dimensional settings. As Eq.~\ref{eq:elbo} suggests, this approach balances the fit to the observed data and the complexity of the model, leveraging the regularisation provided by the priors. The ELBO is maximised using stochastic gradient-based optimisation \citep{kingma2017adam, Hoffman20131303} to jointly infer the latent class structure and the reinforcement learning dynamics underlying the observed data.

Note that the RL model from Section \ref{subsection2:RL} is also estimated using Variational Inference and its generative process is presented in Appendix A. Both the RL and the proposed LCRL models are implemented in Python and PyTorch \citep{Paszke2019}. In addition, a parameter recovery analysis for both models is provided in Appendix B. This validation step assesses the ability of the estimation procedure to recover known parameters from synthetic data, ensuring that the proposed variational inference approach yields reliable and well-identified estimates.

%% file: Data_unmarked.tex
The proposed LCRL model is tested on a driving simulator data set collected at the University of Sydney \citep{Fayyaz2021}. 103 participants completed 20 choice tasks where they had to choose between a route with a fixed travel time (always 5 minutes long), which we refer to as the reliable route, and a route with an uncertain variable travel time (which they were informed could take anywhere between 2 and 7 minutes), which we refer to as the unreliable route. The experiment was divided into two main sections that were one to two weeks apart. 

In the first section, each participant came to the laboratory at the University of Sydney and made 10 choices in the driving simulator (DS), physically experiencing the travel time. If they chose the reliable route, they drove for exactly 5 minutes. If they chose the unreliable route, their driving time was drawn from a pre-specified distribution of either 2 or 7 minutes, where 2 minutes has 60\% probability and 7 minutes has a probability of 40\% (and hence, the average travel time is 4 minutes). Importantly, this travel time distribution is {\em unknown} to participants, and participants only learned the actual duration of the unreliable route upon completion and can learn the travel time distribution through repeated choices. Draws from the travel time distribution for the unreliable route varied across participants, where some received initially a low travel time (2 minutes), while others initially received a high travel time (7 minutes).

In the second section, each participant completed an online questionnaire with 10 stated preference (SP) choice tasks between the same reliable and unreliable routes, where they received immediate written on-screen feedback on the actual duration of their chosen route without physically driving it. 

Participants were randomly assigned to either start with the driving simulator tasks or SP tasks to mitigate ordering effects. Some participants selected the same route across all scenarios and as such were removed from the dataset. The final dataset consists of 1,660 choice observations from 83 individuals. Among these participants, around 49\% are female, 52\% are under 40 years old, 59\% have a full-time job, and 47\% started with SP tasks (Table \ref{T2}).

\begin{figure}[H]
    \centering
    \includegraphics[origin=c, width = 1.0 \textwidth
   ]{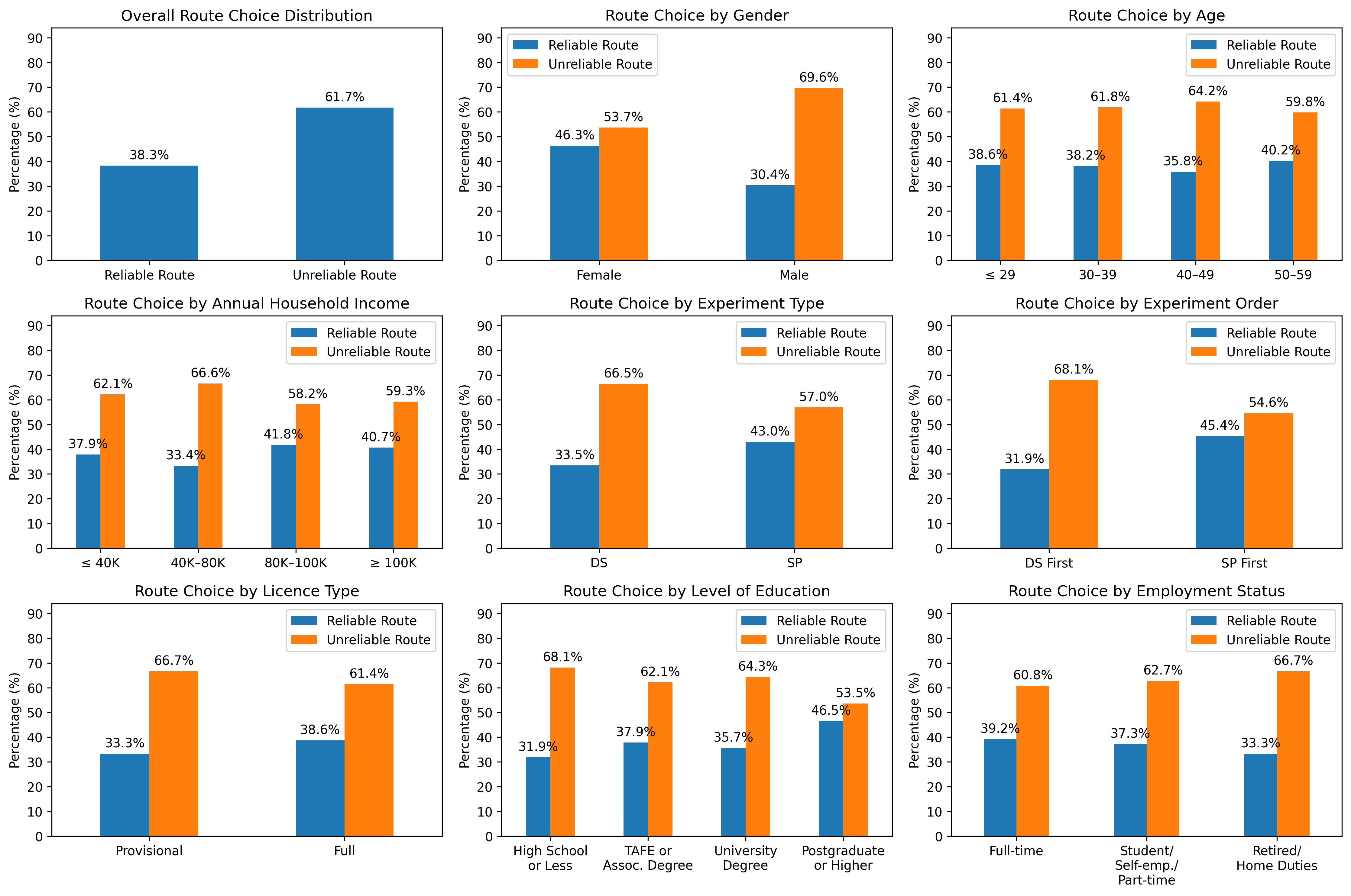}
    \caption{Route choice distribution across demographic and experimental characteristics}
    \label{fig:choices}
\end{figure}

Figure \ref{fig:choices} shows the proportions of reliable versus unreliable route choices across the entire sample and within key socio-demographic/economic and experimental characteristics. Overall, participants chose the unreliable route (61.7\%) more often than the reliable route (38.3\%). Females selected the reliable route in 46.34\% of their choices, more often than males, who chose it only 30.4\% of the time. The preference for the unreliable route thus appeared stronger among males (69.6\%) than females (53.7\%). Across age groups, route preference did not display strong variability, though participants aged 50–59 showed the highest inclination (40.24\%) towards the reliable route compared to other age groups. In terms of income levels, participants from higher income households (AUD 80,000 or more) selected the reliable route slightly more frequently compared to lower income groups. As for education level, participants with a postgraduate degree or higher selected the unreliable route (53.5\%) less often than those from lower education groups (more than 60\%). The experimental context significantly influenced route choice. Participants were more inclined to select the unreliable route in the driving simulator tasks (66.5\%) than in the SP tasks (57\%). Furthermore, those who encountered the SP tasks first were more inclined to select the reliable option (45.4\%) relative to those who began with the DS tasks (31.9\%). Finally, employment status and licence type did not show strong impacts on choice variability across groups.

\begin{figure}[H]
    \centering
    \includegraphics[origin=c, width = 0.6 \textwidth
   ]{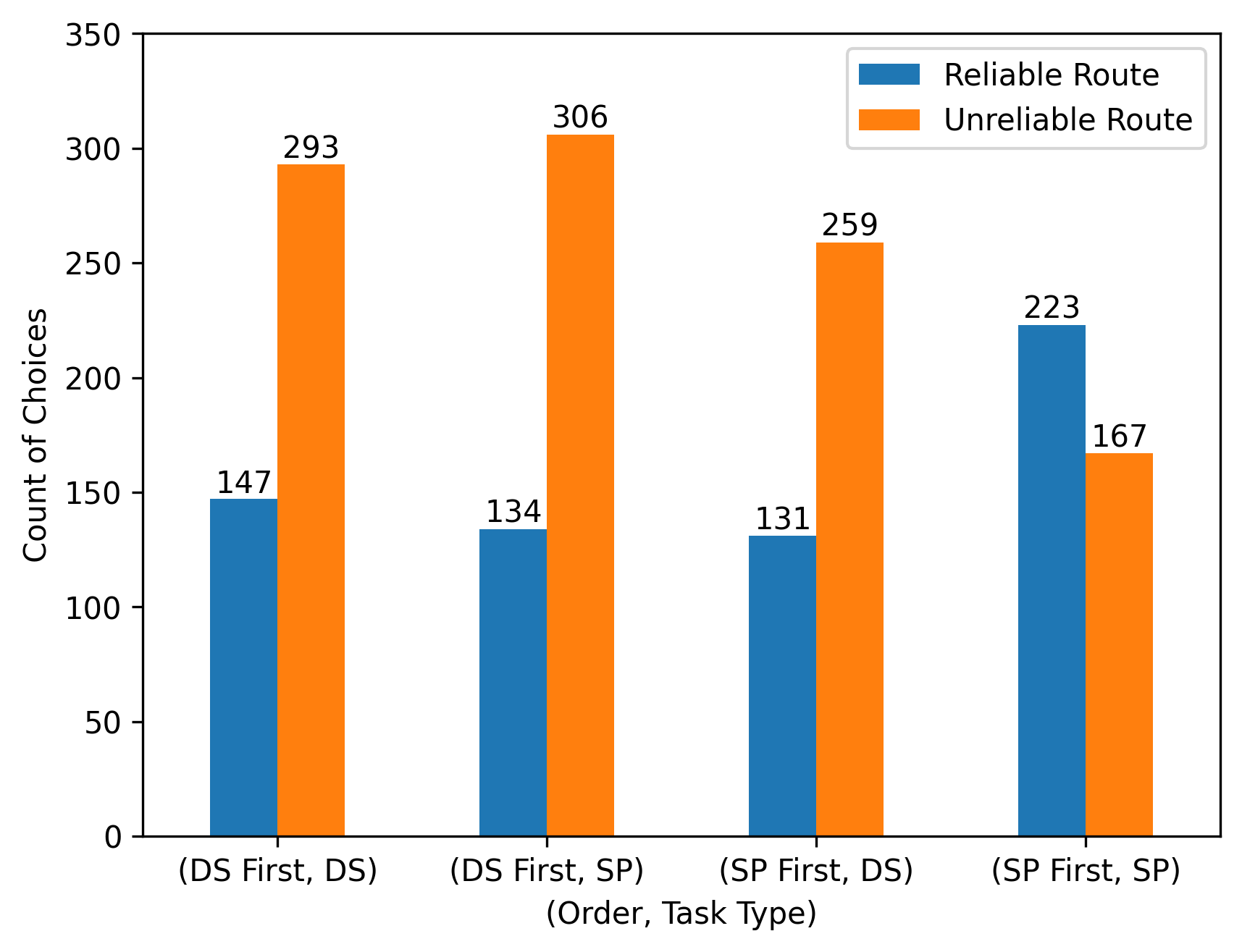}
    \caption{Choice counts by order (DS first vs SP first) and task type (DS vs SP)}
    \label{fig:DSvsSP}
\end{figure}

Figure \ref{fig:DSvsSP} provides additional insights by presenting route choice counts (reliable vs. unreliable), further segmented by both the order of experimental conditions (DS first vs.\ SP first) and the task type (DS vs.\ SP tasks). Participants who started with the driving simulator tasks consistently preferred the unreliable route across both DS (293 unreliable route vs.\ 147 reliable route choices) and subsequent SP tasks (306 unreliable route vs.\ 134 reliable route choices), suggesting that initial exposure to experiential feedback significantly increased participants' risk tolerance. In contrast, participants who started with SP tasks showed an initial preference for the reliable route (223 reliable route vs. 167 unreliable route choices). However, when these same participants later drove the driving simulator, their preference shifted towards the unreliable route (259 unreliable route vs.\ 131 reliable route choices), highlighting the significant role of physical feedback in shaping route choices.

\begin{table}[ht!]
\centering
\caption{Number of participants and route choices across demographic and experimental characteristics}
\label{T2}
\resizebox{\textwidth}{!}{%
\begin{tabular}{l l r r r r}
\hline
\textbf{} & \textbf{} & \multicolumn{2}{c}{\textbf{Number of Participants}} & \multicolumn{2}{c}{\textbf{Number of Choices}} \\
\hline
\textbf{} & \textbf{} & \textbf{Count} & \textbf{\%} & \textbf{Reliable Route} & \textbf{Unreliable Route} \\
\hline
\textbf{Overall Count} & --- & 83 & --- & 635 & 1025 \\
\hline
\multirow{2}{*}{\textbf{Gender}} 
 & Female & 41 & 49.40\% & 380 & 440 \\
 & Male & 42 & 50.60\% & 255 & 585 \\
\hline
\multirow{4}{*}{\textbf{Age}} 
 & 1: 29 or less & 21 & 25.30\% & 162 & 258 \\
 & 2: 30 to 39 & 22 & 26.51\% & 168 & 272 \\
 & 3: 40 to 49 & 19 & 22.89\% & 136 & 244 \\
 & 4: 50 to 59 & 21 & 25.30\% & 169 & 251 \\
\hline
\multirow{4}{*}{\textbf{Income}}
 & 1: 40,000 or less & 7  & 8.43\%  & 53  & 87  \\
 & 2: 40,001 to 80,000 & 28 & 33.73\% & 187 & 373 \\
 & 3: 80,001 to 100,000 & 19 & 22.89\% & 159 & 221 \\
 & 4: More than 100,000 & 29 & 34.94\% & 236 & 344 \\
\hline
\multirow{2}{*}{\textbf{Experiment Type}}
 & Driving Simulator (DS) & 83 & 100.00\% & 278 & 552 \\
 & Stated Preference (SP) & 83 & 100.00\% & 357 & 473 \\
\hline
\multirow{2}{*}{\textbf{Experiment Order}}
 & SP First  & 39 & 46.99\% & 354 & 426 \\
 & DS First  & 44 & 53.01\% & 281 & 599 \\
\hline
\multirow{2}{*}{\textbf{Licence Type}}
 & 2: Provisional   & 6  & 7.23\%  & 40  & 80  \\
 & 3: Full & 77 & 92.77\% & 595 & 945 \\
\hline
\multirow{4}{*}{\textbf{Education Level}}
 & 1: High School or Less & 8  & 9.64\%  & 51  & 109 \\
 & 2: Associate Degree, Diploma & 28 & 33.73\% & 212 & 348 \\
 & 3: University Degree & 30 & 36.14\% & 214 & 386 \\
 & 4: Postgrad or Higher & 17 & 20.48\% & 158 & 182 \\
\hline
\multirow{3}{*}{\textbf{Employment Type}}
 & 1: Full Time & 49 & 59.04\% & 384 & 596 \\
 & 2: Student, Part-Time, Self-Employed & 31 & 37.35\% & 231 & 389 \\
 & 3: Retired or Home Duties & 3  & 3.61\%  & 20  & 40  \\
\hline
\end{tabular}
} 
\end{table}

%% file: Results_unmarked.tex
Given the characteristics of the data presented in the previous section, the Latent Class Reinforcement Learning model introduced in Section \ref{subsection2:LCRL} is adapted accordingly to reflect the participants’ specific decision environment and effectively capture diverse learning dynamics and initial route-choice expectations. Participants were informed that the reliable route would consistently take 5 minutes. Thus, the initial expectation for the reliable route, $Q_{Rel,0,k}$, was set at 5 minutes for all individuals and classes. Conversely, participants were told that the unreliable route's travel time would vary between 2 and 7 minutes. Therefore, the initial expectation for the unreliable route, $Q_{Unr,0,k}$, was modelled probabilistically using the sigmoid transformation outlined in step 2.(c).ii of the generative process in Section \ref{subsection2:LCRL}, with the range boundaries corresponding to the parameters $a=2$ and $b=7$, ensuring that $Q_{Unr,0,k}$ falls within this range.

Furthermore, the sensitivity parameter $\beta_{k}$ is revised to better reflect the empirical characteristics of the dataset. The descriptive analysis (Section \ref{section:dataset}) showed that participants made different route choices during DS and SP tasks as they tended to be more exploratory or risk-seeking in DS tasks (Figure \ref{fig:choices}). To capture that participants' exploration-exploitation behaviour may differ substantially depending on whether they are making choices under actual experienced conditions or hypothetical scenarios, separate class-specific sensitivity parameters for the driving simulator ($\beta_{DS,k}$) and stated preference tasks ($\beta_{SP,k}$) are introduced. Specifically, the sensitivity parameter is defined as follows:
\begin{equation}
    \beta_{k} = \beta_{DS,k} D + \beta_{SP,k} (1-D),
\end{equation}
where $D$ is a dummy variable equal to 1 if the observation corresponds to a driving simulator and 0 otherwise.

Similarly, the bias parameter $\gamma_{ik}$ is defined as follows to capture the influence of the empirical characteristics of the dataset:
\begin{equation}
    \gamma_{ik} = \gamma_{DS,i,k} + \gamma_{SP,i,k} (1 - D),
    \label{eq:gamma_updated}
\end{equation}
where $\gamma_{DS,i,k}$ represents the baseline bias/preference for alternative $i$ in the driving simulator experiments and $\gamma_{SP,i,k}$ is a shift parameter that captures deviations from the DS baseline in the stated preference context.

Based on the descriptive analysis from Section \ref{section:dataset}, we specify the class membership model as a function of key socio-economic/demographic and experimental characteristics. Specifically, the class membership is influenced by gender, age (coded 1 if under 40 years old and 0 otherwise), income (coded 1 if annual household income is below 80,000 AUD and 0 otherwise), education level (coded 1 if participant holds a postgraduate degree or higher and 0 otherwise), and experiment order (whether participants encountered the DS or SP tasks first), in addition to a constant term. 

LCRL models with up to four classes were estimated, along with a standard Reinforcement Learning (RL) model (as introduced in Section \ref{subsection2:RL}), a standard binary logit model (BL) (i.e. a constant only), and a standard latent class choice model (LCCM) without reinforcement learning serving as benchmarks. The same adaptations regarding initial expectations, bias parameters and sensitivity parameters are incorporated into the standard RL model to ensure a fair and consistent comparison. As for the LCCM benchmark, it uses the same class membership specification as the proposed LCRL model, but its class-specific choice model includes only context-dependent route preference terms. Specifically, the benchmark LCCM includes only the class-specific bias parameters $\gamma_{ik}$, specified in Eq. \ref{eq:gamma_updated}, as in the proposed LCRL, but does not include feedback, prediction errors, updated expectations, initial Q-values, learning rates, or exploration-exploitation parameters. It therefore provides a pure latent class discrete choice benchmark against which the contribution of the RL component can be assessed.

Table \ref{tab:model_comparison} reports the goodness-of-fit statistics for the BL, RL, LCCM, and LCRL models. The single-class benchmarks, BL and RL (K = 1), perform worst on every criterion, yielding the lowest log-likelihood (LL) and the highest Akaike information criterion (AIC) and Bayesian information criterion (BIC) values. The BIC is calculated using the number of independent individuals, $N=83$.

The single-class BL model, with a single parameter, fits substantially worse than RL, indicating that the learning component captures behaviour that a constant-utility specification cannot. Introducing latent classes substantially improves model fit, as shown by the improvements obtained by both the LCCM and LCRL specifications relative to the single-class BL and RL models. The two-class LCCM raises the LL to -913.96 and the three-class LCCM improves it further to -848.03, confirming that static latent preference heterogeneity and DS/SP context effects explain an important part of the observed behaviour. For RL, there is an improvement for moving from a single-class to a two-class model (-962.91 to -870.16) while the three-class LCRL further improves the LL to -803.51. 
The four-class LCRL achieves the best LL and AIC, but its BIC of 1,758.05 is higher than that of the three-class LCRL. Across specifications with the same number of classes, the RL models always outperform the models without a learning component. The three-class LCRL has the lowest BIC of all specifications considered, at 1,739.59, compared with 1,775.60 for the three-class LCCM. Accordingly, the three-class LCRL is retained as the preferred specification, as it provides the best balance between model fit and parsimony.

\begin{table}[h!]
\centering
\caption{Model comparison statistics}
\label{tab:model_comparison}
\begin{tabular}{c c c c c c}
\hline
Model & K & Nb. Param & LL & AIC & BIC\\
\hline
BL   & 1 & 1  & -1,104.38 & 2,210.76 & 2,213.18 \\
RL   & 1 & 6  & -962.91 & 1,937.82 & 1,952.33 \\
\hline
LCCM & 2 & 10 & -913.96 & 1,847.92 & 1,872.11 \\
LCCM & 3 & 18 & -848.03 & 1,732.06 & 1,775.60\\
\hline
LCRL & 2 & 18 & -870.16 & 1,776.32 & 1,819.86\\
LCRL & 3 & 30 & -803.51 & 1,667.02 & \textbf{1,739.59}\\
LCRL & 4 & 42 & \textbf{-786.23} & \textbf{1,656.46} & 1,758.05\\
\hline
\end{tabular}
\end{table}

\begin{table}[ht!]
\centering
\caption{RL, LCCM and LCRL results}
\label{tab:param_estim}
\resizebox{\textwidth}{!}{
\begin{tabular}{l c c c c c c c}
\toprule
&
\multicolumn{1}{c}{\textbf{RL}} & \multicolumn{3}{c}{\textbf{LCCM}} & \multicolumn{3}{c}{\textbf{LCRL}} \\
\cmidrule(lr){2-2} \cmidrule(lr){3-5} \cmidrule(lr){6-8} \textbf{Variable} & & \textbf{Class 1} & \textbf{Class 2} & \textbf{Class 3} & \textbf{Class 1} & \textbf{Class 2} & \textbf{Class 3}
\\
\midrule
& \textbf{RL Choice Model} & \multicolumn{3}{c}{\textbf{LCCM Benchmark}} & \multicolumn{3}{c}{\textbf{Class-Specific RL Choice Model}}
\\
\cmidrule(lr){2-2} \cmidrule(lr){3-5} \cmidrule(lr){6-8}

$\gamma_{\mathrm{DS},Rel,k}$ & $-0.799\,(-14.81)$ & $-0.270\,(-1.89)$ & $-1.61\,(-15.84)$ & $0.0675\,(0.77)$ & $-0.635\,(-4.00)$ & $-1.49\,(-15.87)$ & $0.372\,(3.88)$\\

$\gamma_{\mathrm{SP},Rel,k}$ & $0.299\,(2.77)$ & $3.92\,(9.77)$ & $-0.0655\,(-0.47)$ & $-0.342\,(-2.91)$ & $3.70\,(8.37)$ & $0.390\,(2.95)$ & $-0.684\,(-5.17)$\\

$\beta_{\mathrm{DS},k}$ & $0.419\,(5.81)$ & -- & -- & -- & $0.935\,(0.40)$ & $0.337\,(3.42)$ & $0.247\,(4.27)$\\

$\beta_{\mathrm{SP},k}$ & $1.00\,(0.03)$ & -- & -- & -- & $0.781\,(0.62)$ & $0.837\,(1.11)$ & $0.141\,(4.55)$\\

$\alpha_k$ & $0.251\,(8.32)$ & -- & -- & -- & $0.277\,(3.55)$ & $0.355\,(2.19)$ & $0.437\,(0.42)$\\

$Q_{\mathrm{Unr},0,k}$ & $6.69\,(8.25)$ & -- & -- & -- & $6.16\,(5.31)$ & $6.12\,(4.05)$ & $4.81\,(0.36)$\\

\midrule

& & \multicolumn{3}{c}{\textbf{Class Membership}} & \multicolumn{3}{c}{\textbf{Class Membership}}\\

\cmidrule(lr){3-5}
\cmidrule(lr){6-8}

$\eta_{\mathrm{ASC}}$ & & $-0.108\,(-0.22)$ & $-0.184\,(-0.40)$ & & $-0.0403\,(-0.08)$ & $0.0291\,(0.06)$ & \\

$\eta_{\mathrm{DS\,First}}$ & & $-1.00\,(-2.07)$ & $1.14\,(3.08)$ & & $-1.02\,(-2.10)$ & $0.943\,(2.53)$ & \\

$\eta_{\mathrm{Female}}$ & & $-0.0329\,(-0.07)$ & $-1.21\,(-2.88)$ & & $-0.121\,(-0.27)$ & $-1.36\,(-3.13)$ & \\

$\eta_{\mathrm{Age<40}}$ & & $0.0256\,(0.06)$ & $0.0945\,(0.24)$ & & $0.170\,(0.38)$ & $0.297\,(0.75)$ & \\

$\eta_{\mathrm{Income<80k}}$ & & $-0.322\,(-0.74)$ & $0.812\,(2.27)$ & & $-0.222\,(-0.51)$ & $0.984\,(2.75)$ & \\

$\eta_{\mathrm{PostGrad}}$ & & $0.393\,(0.78)$ & $-0.0365\,(-0.07)$ & & $0.408\,(0.81)$ & $0.0849\,(0.17)$ & \\

\midrule

Class Share & & $21.57\%$ & $45.41\%$ & $33.02\%$ & $21.59\%$ & $49.10\%$ & $29.31\%$\\

LL & $-962.91$ & \multicolumn{3}{c}{$-848.03$} & \multicolumn{3}{c}{$-803.51$}\\

AIC & $1{,}937.82$ & \multicolumn{3}{c}{$1{,}732.06$} & \multicolumn{3}{c}{$1{,}667.02$}\\

BIC & $1{,}952.33$ & \multicolumn{3}{c}{$1{,}775.60$} & \multicolumn{3}{c}{$1{,}739.59$}\\

Nb. Param & $6$ & \multicolumn{3}{c}{$18$} & \multicolumn{3}{c}{$30$}\\

\bottomrule
\end{tabular}
}
\end{table}


Table \ref{tab:param_estim} presents the results of the standard RL, three-class LCCM, and three-class LCRL models. For each parameter, we report the posterior mean and an approximate posterior z-score, in parentheses, defined as the ratio of the posterior mean to the posterior standard deviation. As previously mentioned, the three-class LCRL model outperforms the RL benchmark in terms of LL, AIC, and BIC. It also outperforms the three-class LCCM across all three criteria. These results indicate that both components of the proposed framework contribute to explaining the observed choices. The comparison between the RL and LCCM benchmarks shows that class-specific, context-dependent route preferences account for a substantial part of the observed heterogeneity. At the same time, the improvement of the LCRL over the LCCM indicates that the reinforcement learning component provides additional explanatory power beyond a static latent class structure. The LCRL model identifies three distinct latent classes, which can be labelled and described as follows. Note that given the stylised nature of the empirical setting and the modest sample size, these classes should be interpreted as behavioural patterns identified within the present route choice experiment rather than as definitive population wide traveller typologies.
\\\\
\textit{Class 1 - SP-first participants with context-dependent preferences and exploitative tendencies}: The first class represents approximately 22\% of the sample. Participants who encountered the SP tasks first have a higher posterior probability of belonging to this class (`Driving Simulator First' coefficient is negative and significant). The remaining class membership coefficients are insignificant. Members of this class display strong context-dependent preferences. They show a statistically significant aversion to the reliable route in the driving simulator ($\gamma_{DS,Rel,k}$ = -0.635 and significant), suggesting a preference for the unreliable route when outcomes are physically experienced. However, the large positive and significant SP-shift parameter ($\gamma_{SP,Rel,k}$ = 3.70) indicates a clear reversal of this preference under hypothetical SP conditions, where participants become clearly more inclined towards the reliable route. Both sensitivity parameters are positive ($\beta_{DS,k}$ = 0.935 and $\beta_{SP,k}$ = 0.781) but statistically insignificant, indicating that choices are not strongly governed by the exploration–exploitation mechanism captured by the sensitivity parameters. Instead, the consistent selection of different routes across contexts reflects context-specific exploitation, driven by shifts in the bias parameters rather than by systematic exploitation in the RL sense. Finally, the learning rate is low ($\alpha_{k}$ = 0.277) but significant, indicating that these participants are slow learners who update their expectations slowly while relying on older experiences to make choices. \\\\
\textit{Class 2 - DS-first participants with persistent exploitative tendencies}: The second class accounts for nearly half of the participants (49.10\%) and mainly includes male participants with low income who encountered the driving simulator tasks first, as indicated by statistically significant coefficients associated with these variables. Participants in this class exhibit a strong and highly significant preference for the unreliable route in the driving simulator ($\gamma_{DS,Rel,k}$ = -1.49), more than twice the magnitude observed in class 1. The small positive and significant SP-shift parameter ($\gamma_{SP,Rel,k}$ = 0.390) suggests a slightly lower aversion to the reliable alternative in hypothetical tasks but without a fundamental change in preferences. Unlike the first class, the DS sensitivity parameter ($\beta_{DS,k}$ = 0.337) is significant, indicating an exploitative behaviour when outcomes are physically experienced, while the SP sensitivity parameter ($\beta_{SP,k}$ = 0.837) is larger but statistically insignificant. The learning rate is moderate ($\alpha_{k}$ = 0.355), implying a somewhat faster adaptation to feedback than Class 1 participants, though these participants still rely considerably on past experiences when making decisions. \\\\
\textit{Class 3 - Participants with context-dependent preferences and exploratory tendencies}: Approximately 29\% of the sample belong to this class, which is more likely to include females and higher-income participants. Unlike the other two classes, members of Class 3 have a significant preference for the reliable route in the driving simulator ($\gamma_{DS,Rel,k}$ = 0.372), reflecting a clear reliability-seeking tendency. However, the negative and significant SP-shift parameter ($\gamma_{SP,Rel,k}$ = -0.684) reverses this preference in the hypothetical setting. The resulting reliable route bias in the SP task is $0.372 - 0.684 = -0.312$, indicating a greater inclination towards the unreliable route. Both sensitivity parameters are significant but relatively low ($\beta_{DS,k}$ = 0.247 and $\beta_{SP,k}$ = 0.141), indicating more explorative tendencies compared to the other two classes. The learning rate is the highest among the three classes ($\alpha_{k}$ = 0.437) but statistically insignificant, suggesting relatively fast updating of expectations in response to recent feedback, although this estimate is imprecise. Finally, participants in this class display an average initial perceived travel time for the unreliable route ($Q_{Unr,0,k}$ = 4.81) in contrast to Classes 1 and 2 who initially perceived the unreliable route as having relatively high travel time (6.16 and 6.12, respectively), close to the maximum possible travel time of 7 minutes.

It is also useful to compare the latent class structure of the LCRL model with that obtained from the standard LCCM benchmark. The two models produce similar class shares as shown in Table \ref{tab:param_estim}. The class membership parameters are also consistent across the two models in terms of signs and relative magnitudes. This consistency suggests that the LCRL model does not generate a fundamentally different segmentation simply by introducing RL dynamics. Instead, the LCRL preserves the main latent class structure identified by the static LCCM while adding information on how classes differ in their learning rates, initial expectations, and exploration-exploitation sensitivities. Moreover, the class-specific choice parameters of the LCCM also align closely with the bias parameters of the LCRL. In both models, Class 1 displays a strong context-dependent reversal in route preferences, moving from a preference for the unreliable route in the driving simulator to a strong preference for the reliable route in the SP tasks. Class 2 displays a persistent preference for the unreliable route, especially in the driving simulator, with only a modest change in the SP context. Class 3 is comparatively more reliability-oriented in the driving simulator and more inclined towards the unreliable route in the SP tasks. These similarities confirm that context-dependent route preference heterogeneity is an important component of the observed behaviour. However, the LCRL extends this static interpretation by showing that the classes also differ in their feedback-driven learning dynamics, as previously discussed. The comparison with the LCCM therefore indicates that the proposed LCRL captures both static context-dependent preference heterogeneity and additional heterogeneity in feedback-driven learning dynamics.

\begin{figure}[H]   
    \centering
    \includegraphics[origin=c, width = 1.0 \textwidth
   ]{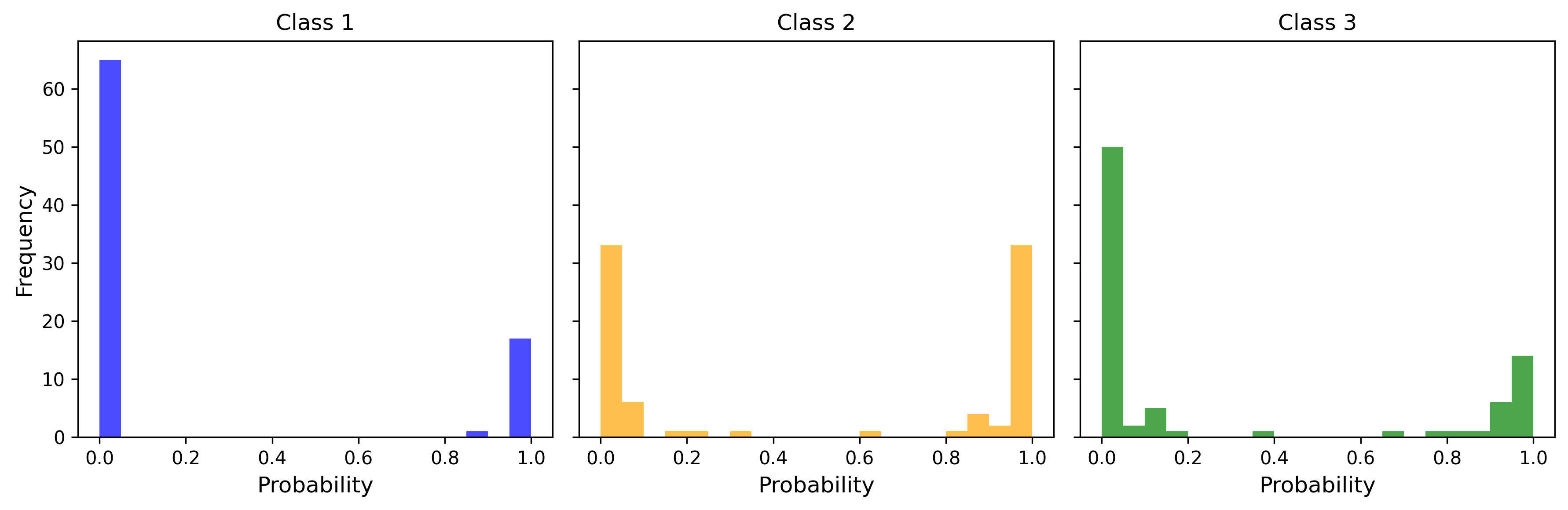}
    \caption{Posterior probabilities of class assignments (class membership probabilities) - LCRL}
    \label{fig:q_s}
\end{figure}

Figure \ref{fig:q_s} shows the posterior class-membership probabilities for each of the three latent classes identified by the LCRL model. The distributions exhibit clear bimodality, with most individuals having probabilities close to either 0 or 1 for each class. This pattern indicates strong posterior certainty as most individuals are assigned to one class with high probability. Such sharp separation confirms that the LCRL model uncovers well-defined and behaviourally distinct latent classes.

\begin{figure}[H]   
    \centering
    \includegraphics[origin=c, width = 1.0 \textwidth
   ]{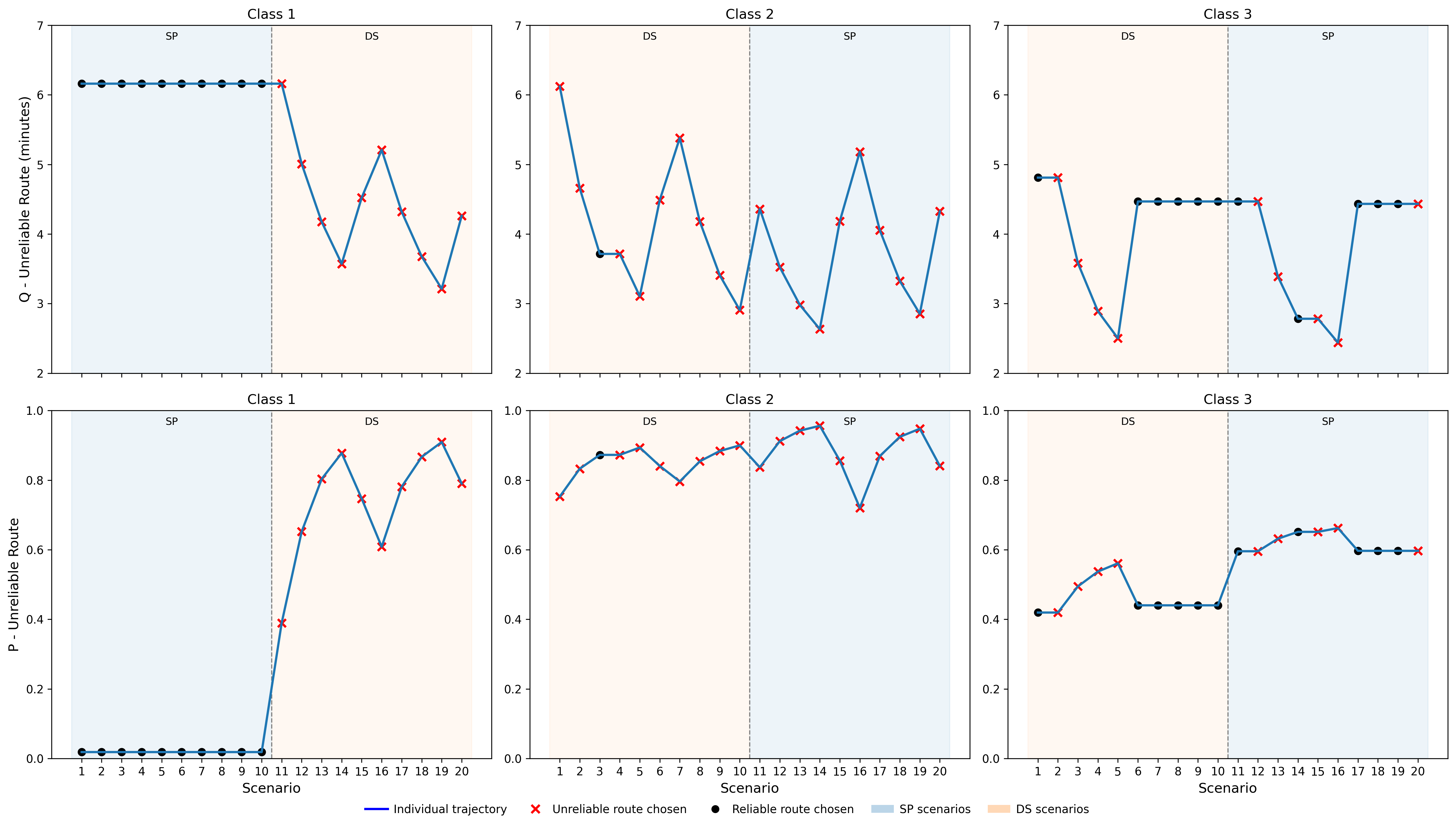}
    \caption{Travel time expectations and choice probabilities for the unreliable route, together with the observed choice trajectories of three individuals, each associated with a distinct class, across 20 scenarios (10 from the driving simulator and 10 from the stated-preference survey)}
    \label{fig:Q_P_choice}
\end{figure}

Moreover, Figure \ref{fig:Q_P_choice} illustrates the travel time expectations of the unreliable alternative, the choice probabilities for this alternative, and the realised choice trajectories of three individuals who each have a posterior class-membership probability greater than 0.9 for one of the three latent classes. Each of these individuals can therefore be regarded as deterministically belonging to their respective classes. The trajectories provide a within-participant visualisation of how the class-specific parameters in Table \ref{tab:param_estim} translate into behavioural patterns across the 20 decision scenarios.

The individual belonging to Class 1 selects the reliable route in all SP scenarios, with a near-zero probability of choosing the unreliable route. This behaviour is consistent with the large positive and significant SP-shift parameter ($\gamma_{SP,Rel,k}$ = 3.70), which strongly favours the reliable route in hypothetical scenarios. When switching to the DS experiment, this preference reverses as the probability of choosing the unreliable route increases sharply, and the individual selects this option throughout the DS scenarios regardless of the feedback experienced. This aligns with the negative and significant DS bias parameter ($\gamma_{DS,Rel,k}$ = -0.635), which indicates a clear preference for the unreliable route when outcomes are physically experienced in the driving simulator. Expectations adjust only gradually during the DS tasks, reflecting the slow learning rate estimated for this class. 

The individual representing Class 2 exhibits a different pattern. Across both DS and SP scenarios, this participant chooses the unreliable route in all scenarios, except only once (in scenario 3), with high probabilities that remain above 0.7 throughout. This persistent preference accords with the strong negative DS bias ($\gamma_{DS,Rel,k}$ = -1.49) and the small but positive SP-shift parameter ($\gamma_{SP,Rel,k}$ = 0.390), both of which favour the unreliable route overall. This exploitative behaviour of the unreliable route is also consistent with the significant DS sensitivity parameter ($\beta_{DS,k}$ = 0.337) and the relatively high, though insignificant, SP sensitivity parameter ($\beta_{SP,k}$ = 0.837). Compared with Class 1, the travel time expectations fluctuate more sharply, consistent with the moderate and significant learning rate ($\alpha_{k}$ = 0.355), which leads to faster adjustments in response to experienced feedback. 

In contrast, the individual assigned to Class 3 displays more exploratory behaviour, switching between the two alternatives across both contexts. This aligns with the relatively low sensitivity parameters ($\beta_{DS,k}$ = 0.247 and $\beta_{SP,k}$ = 0.141), which lead to more exploratory choice patterns. Nonetheless, this individual shows a tendency to favour the reliable route in the DS scenarios (selecting it 6 out of 10 times), in accordance with the positive DS bias ($\gamma_{DS,Rel,k}$ = 0.372). In SP settings, this preference shifts towards the unreliable route (selecting it 6 out of 10 times), consistent with the negative and significant SP-shift parameter ($\gamma_{SP,Rel,k}$ = -0.684).


\begin{figure}[H]   
    \centering
    \includegraphics[origin=c, width = 1.0 \textwidth
   ]{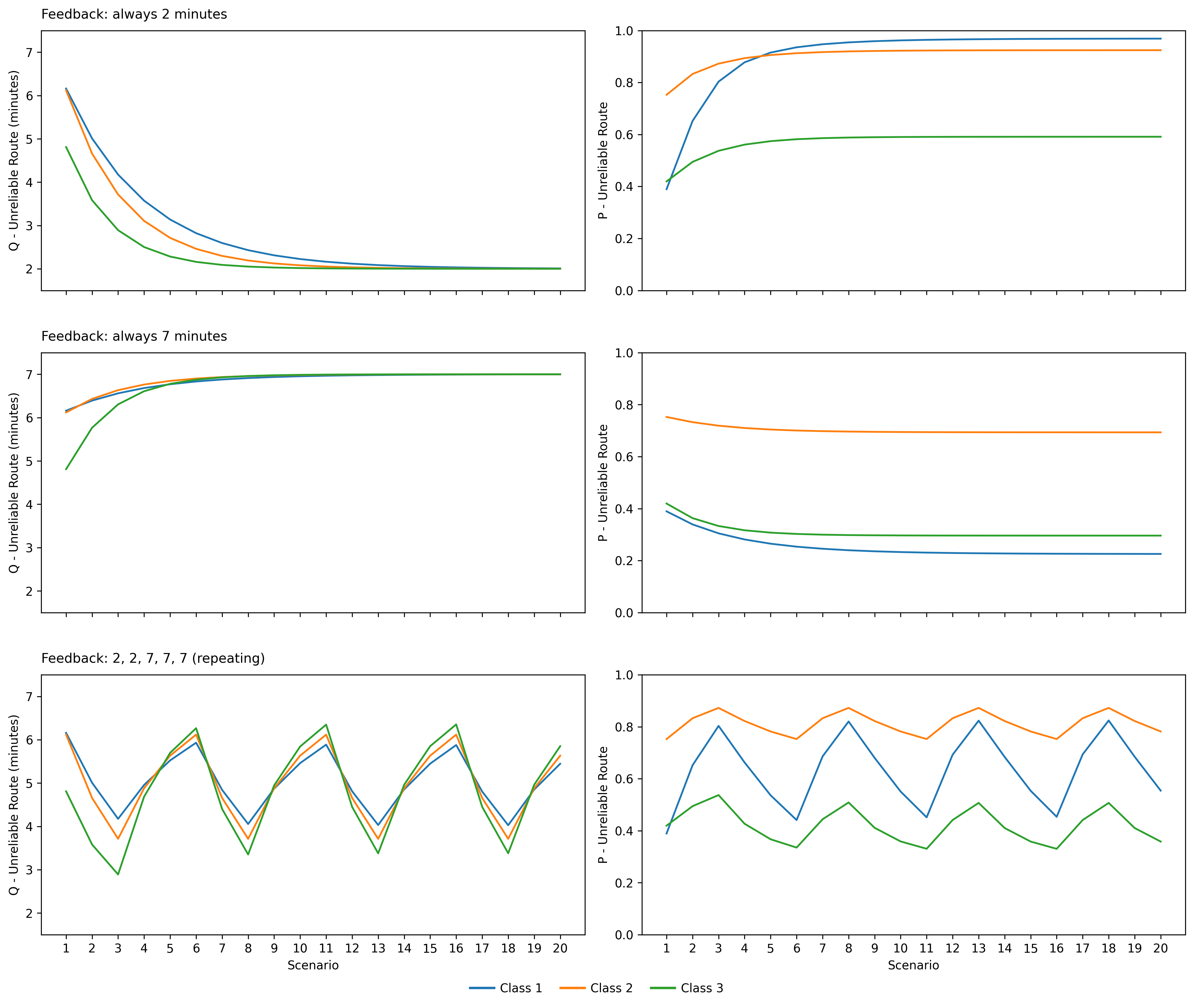}
    \caption{Expectation and probability trajectories by class - DS scenarios}
    \label{fig:Q_P_sim_DS}
\end{figure}

Next, Figure \ref{fig:Q_P_sim_DS} presents simulated trajectories of both the travel time expectations and the choice probabilities, of one representative individual from each class, for the unreliable route under three “extreme” experiments. In all experiments, an individual is assumed to always select the unreliable route across 20 driving-simulator scenarios while receiving one of the following feedback patterns: a) a constant travel time of 2 minutes, b) a constant travel time of 7 minutes, or c) a repeating sequence of 2, 2, 7, 7, 7 minutes. These simulated cases highlight how the class-specific learning rates and sensitivity parameters shape the evolution and convergence of expectations and choice probabilities under conditions of persistent (always 2 or always 7 minutes) or alternating feedback. Consistent with the estimates reported in Table \ref{tab:param_estim}, Class 3 exhibits the fastest convergence of expectations under stable feedback, reflecting its comparatively high learning rate. Under the alternating feedback scenario, the three classes display different repetitive patterns, with Class 3 showing the strongest and most rapid expectation updating, again consistent with its high learning rate compared to the other two classes. Regarding choice probabilities, Class 2 consistently shows high values across all feedback scenarios, reflecting its strong preference for the unreliable route. Moreover, Class 2 displays minimal variation around 0.8 regardless of the feedback received, indicating its exploitative tendencies. In contrast, the choice probability of Class 3 oscillates around 0.4, reflecting its more reliability-seeking behaviour in driving-simulator settings.

\begin{figure}[H]   
    \centering
    \includegraphics[origin=c, width = 1.0 \textwidth
   ]{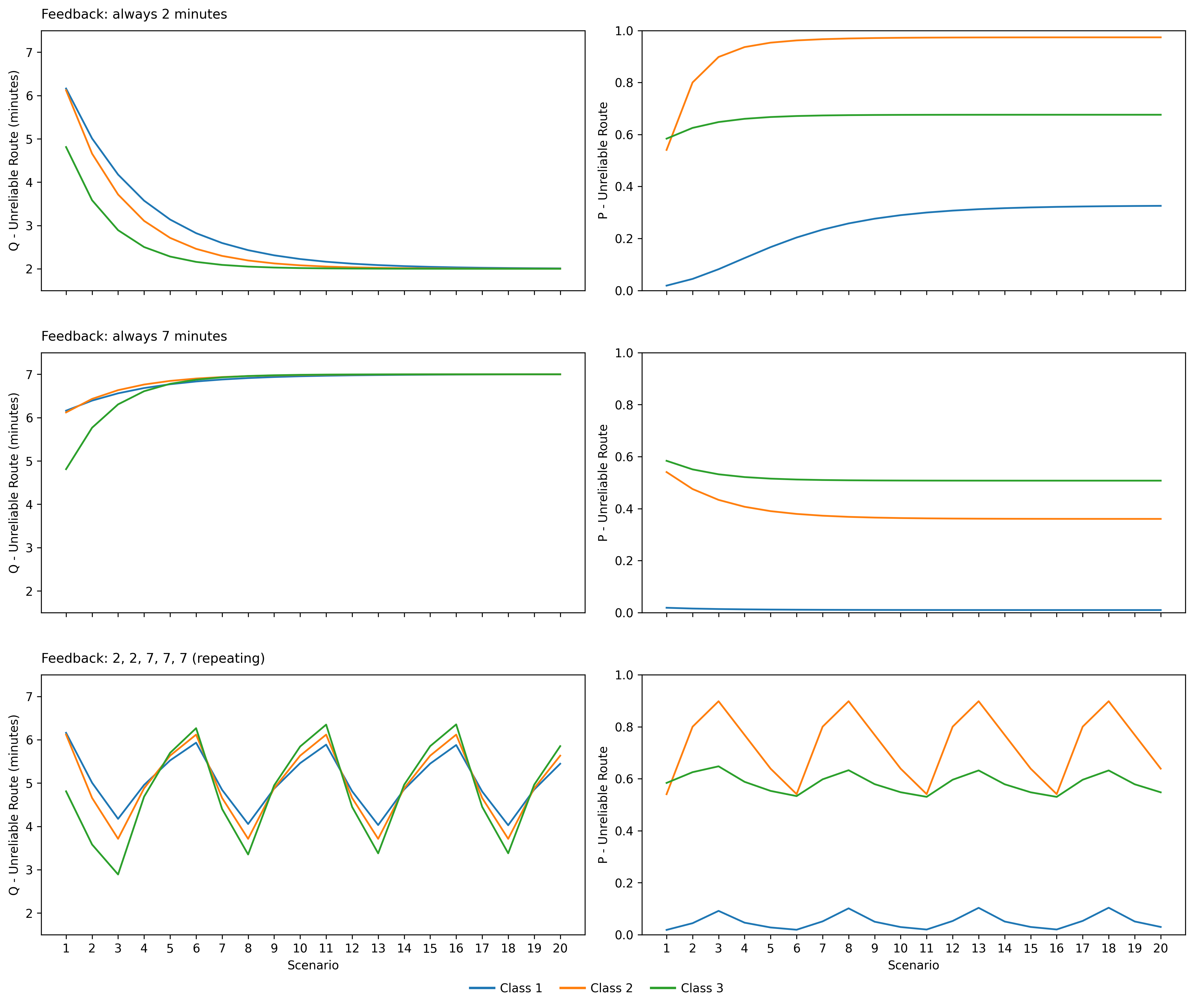}
    \caption{Expectation and probability trajectories by class - SP scenarios}
    \label{fig:Q_P_sim_SP}
\end{figure}

Similarly, Figure \ref{fig:Q_P_sim_SP} presents the same simulated experiments under SP conditions. While Class 1 showed on average high choice probabilities for the unreliable route in DS settings (Figure \ref{fig:Q_P_sim_DS}), Figure \ref{fig:Q_P_sim_SP} shows that this class exhibits low choice probabilities for the unreliable route in all SP scenarios, regardless of the feedback received. This reversal reflects the context-dependent preference of this group (strongly favouring the unreliable option in DS settings, yet preferring the reliable option in hypothetical SP settings). For Class 3, the choice probabilities vary in a pattern similar to that observed in DS conditions, but shift upwards and centre around a higher value of approximately 0.6 (compared to 0.4 in DS). This upward shift reflects its more risk-seeking tendencies in hypothetical SP settings.

Overall, these simulated trajectories emphasise the behavioural differences uncovered by the LCRL model and provide further intuition into how each class responds to sustained or fluctuating experience in extreme learning environments.

Moreover, the proposed framework is also evaluated through a synthetic validation and parameter recovery study, reported in Appendix B. This exercise is designed to complement the empirical application by assessing whether the LCRL model and the variational inference procedure can recover known data-generating parameters under controlled conditions. The synthetic datasets are generated to replicate the core structure of the real data from Section \ref{section:dataset} and the two-class model specification used in this section, including the same number of respondents, repeated choice tasks, driving simulator and stated preference contexts, route travel-time structure, context-specific sensitivity and bias parameters, learning rates, initial expectations, and class-membership parameters. The results show that the main behavioural and learning parameters are reliably recovered, with somewhat greater variability for class-membership parameters, as expected when simulated classes are less behaviourally separated. The synthetic validation therefore provides evidence that the proposed framework can recover the main data-generating parameters under a controlled design mirroring the empirical application, reducing the concern that the empirical patterns reported in this section are simply artefacts of the estimation procedure.

We additionally assessed the stability of the preferred three-class LCRL specification by re-estimating the model 15 times using different random initialisations. After aligning the class labels across runs, the resulting solutions showed similar model fit, with log-likelihood values ranging from -802.68 to -818.41. Ten of the fifteen runs were within 1\% of the log-likelihood reported in Table \ref{tab:model_comparison}, and all were within 1.85\%. The aligned class-specific reinforcement learning parameters differed by no more than 6\%, and the qualitative behavioural interpretation of the three classes remained unchanged across runs. Together with the concentrated posterior class-membership probabilities in Figure \ref{fig:q_s} and the synthetic parameter recovery results in Appendix B, these findings provide evidence of optimisation stability, posterior class separation, and parameter recoverability. These checks should nevertheless be interpreted as complementary internal validation rather than as a substitute for external validation using an independent sample.

Finally, and as with all variational inference approaches, the resulting posterior distributions reported in this section should be interpreted as approximations to the exact Bayesian posterior. In the present model, the mean-field factorisation improves computational tractability but does not capture posterior dependence between parameters and may underestimate posterior uncertainty.

%% file: Conclusions_unmarked.tex
This study bridges econometric discrete choice models with mathematical psychology by introducing a fully Bayesian Latent Class Reinforcement Learning model, estimated through variational inference techniques from machine learning. Similar to traditional Latent Class Choice Models, the proposed LCRL model consists of two sub-components: a latent class membership model and a class-specific choice model. The former accounts for unobserved heterogeneity as discrete latent constructs or classes, allowing for representation of distinct learning types rather than assuming a single behavioural process. This structure enables the model to uncover systematic differences across groups in how individuals explore or exploit alternatives, update their beliefs, and make choices, something that standard RL or continuous random-parameter models typically overlook. The latter captures class-specific learning dynamics using the Rescorla-Wagner reinforcement learning model.

Conventional discrete choice models typically capture stable preferences under static conditions, whereas reinforcement learning models focus explicitly on dynamic and adaptive behaviours that evolve through feedback or reward received from past experiences. By integrating these two perspectives, the proposed LCRL model offers a more comprehensive understanding of how individuals update their decision-making preferences dynamically over time. Specifically, reinforcement learning models gain the ability to account for unobserved heterogeneity through latent class segmentation, capturing different learning dynamics across individuals. Conversely, latent class discrete choice models from econometrics benefit from the ability to model dynamic preference evolution and exploration-exploitation trade-off behaviour through class-specific RL choice models. However, the proposed LCRL model should be interpreted as a model of bounded adaptive learning rather than as a complete cognitive account of habit formation or attention. Travellers are not assumed to know the underlying outcome distribution. Instead, they update subjective expectations incrementally from experienced feedback through prediction errors. This captures an important form of bounded rationality and can describe feedback-driven movement towards more stable or apparently more exploitative choice patterns. However, the present model does not explicitly model habit formation in the psychological sense, nor does it represent attentional allocation, cognitive effort, or uncertainty-directed exploration. Extending the framework to incorporate these mechanisms, for example through attention-dependent learning rates, uncertainty-based exploration, or model-based/model-free components, represents a promising direction for future research.

The LCRL model was tested on a route choice dataset collected through driving simulator and stated preference experiments. The model substantially outperformed the benchmark RL model and also improved on the standard LCCM benchmark in terms of LL, AIC, and BIC, which itself outperforms a simple binary logit model, thus also highlighting heterogeneity. These comparisons show that while static, context-dependent preference heterogeneity explains an important part of the observed behaviour, the RL component provides additional explanatory power by capturing feedback-driven expectation updating and class-specific exploration--exploitation tendencies. The empirical results revealed (i) a context-dependent preference group mainly composed of SP-first participants who switch from choosing a reliable route in hypothetical scenarios to choosing an unreliable route when physically experiencing outcomes in a driving simulator; (ii) a strongly exploitative group who consistently favour the unreliable route across both DS and SP contexts and adapt moderately to feedback; and (iii) a more exploratory group, more likely females and higher-income participants, that shows a reliability preference in DS tasks but reverses this preference under SP conditions. These findings demonstrate that individuals differ not only in their context-dependent preferences but also in how they learn, how quickly they update expectations, and how they balance exploration and exploitation. The results also have important implications for travel behaviour modelling, as they help explain discrepancies between stated and experienced choices and offer a richer understanding of how travellers adapt to variability and uncertainty in transport networks.

Beyond the travel behavioural insights offered in this study, the methodological innovation itself constitutes an emerging modelling technology. The integration of econometric choice models with reinforcement learning and Bayesian inference provides a new framework for capturing adaptive, experience-based decision-making. While our empirical application focuses on human route choice, the approach has direct relevance to emerging mobility technologies. For example, this approach may inform the development of autonomous vehicle (AV) routing strategies. As AVs will operate in environments shaped by human decision-makers, the proposed LCRL model could support the design of AV routing algorithms that better anticipate human learning and expectation formation processes \citep{Zhang2025}, ultimately promoting socially aligned and user-acceptable navigation strategies in mixed-autonomy systems. Similarly, the proposed LCRL model can be of interest to other emerging mobility contexts in which learning, adaptation, and heterogeneity are central. In Mobility-as-a-Service (MaaS) platforms, travellers often experiment with unfamiliar modes or bundling options before converging to habitual usage patterns. An LCRL-based approach could complement existing MaaS segmentation studies that rely on static latent class clustering \citep{GAN202557} by introducing dynamic learning behaviour into the analysis and identifying which segments are more likely to explore and how incentives or information might accelerate adoption. Likewise, real-time information and personalised route guidance systems increasingly rely on behavioural prediction modules to shape user interaction. The LCRL framework, by capturing both expectation updating and class-specific exploration-exploitation tendencies, offers a promising tool for designing information strategies that recognise adaptive behaviour rather than assuming static preferences. The model also offers valuable insights for transport planning and policy in situations where travellers must deviate from habitual behaviour. Because it captures how individuals explore new options and update expectations following unexpected outcomes, it can help policymakers anticipate behavioural responses to disruptions such as network closures, incidents, or service unreliability. Likewise, the model can inform assessments of how travellers learn to use newly introduced infrastructure, such as new metro lines, motorway links, or cycling corridors, by identifying which groups are more inclined to experiment early and how quickly different segments converge to new stable patterns. Incorporating these adaptive learning processes can therefore support more accurate demand forecasting and the design of targeted information or incentive strategies during periods of network change. The framework also provides a basis for class-specific policy simulations that examine how different traveller segments adapt over time to changes in information provision, new alternatives, disruptions, or travel time distributions. Extending such analyses to welfare assessment would require additional assumptions linking learned subjective expectations and experienced outcomes to welfare measures and is therefore left for future research.

Finally, the present empirical application should be viewed as a first-step RL specification in which the learned object is a scalar expected travel time. The repeated feedback structure of the experiment means that participants could, in principle, learn not only the expected travel time of the unreliable route, but also aspects of its reliability, such as the variability or frequency of long and short travel-time outcomes, as well as familiarity with routes that are repeatedly chosen. Therefore, these behavioural mechanisms are not necessarily absent from the experience of participants, but they are not modelled explicitly in the current specification. This modelling choice was made to isolate and demonstrate the core feedback-driven learning mechanism within the proposed latent class reinforcement learning framework. Moreover, given the size of the empirical dataset and the already substantial number of class-specific learning, sensitivity, bias, and membership parameters, we deliberately retained a parsimonious model of expected travel time learning. Incorporating separate learning processes for expected travel time, reliability, familiarity, and risk sensitivity would substantially increase the number of latent states and parameters, raising identification and overfitting concerns in the present application. We therefore view these richer specifications as important extensions for future datasets with more repeated observations per individual and/or experimental variation designed to separately identify these mechanisms.

Specifically, the proposed LCRL model could be extended to incorporate these behavioural factors more explicitly. Reliability perception could be represented through a richer learning structure in which individuals update both the expected travel time and the uncertainty or variability associated with each route, for example using a mean–variance RL model or a Bayesian/Kalman-style learner. Familiarity or habit could be introduced through a perseveration or choice-kernel term that increases with previous choices of the same route and enters utility alongside the learned travel time expectation. Risk attitudes could be modelled through a risk-sensitive value function, allowing travellers to respond not only to the mean outcome but also to variability, downside risk, or asymmetric responses to unexpectedly good and bad outcomes. Route complexity is more directly linked to the experimental design, because the present application uses a deliberately simple binary route choice task. Future applications could incorporate complexity by varying the number of alternatives, information load, route overlap, or network structure, and by allowing these features to affect either the value-learning process or the choice sensitivity parameter. Future research could also explore alternative representations of heterogeneity, including hierarchical RL models with continuous individual-level variation in learning and sensitivity parameters \citep{VANGEEN2021102602}, or hybrid discrete-continuous specifications that allow within-class heterogeneity \citep{Greene01052013}. A systematic comparison between discrete and continuous representations of heterogeneity is beyond the scope of the present study, as the preferred structure is likely to depend on the data, the behavioural question, and/or the modeller’s objective.


%% file: Acknowledgment_unmarked.tex
The authors acknowledge the financial support by the European Research Council through the advanced grant 101020940-SYNERGY and the Australian Research Council through Discovery Project DP150103299. The first author used AI tools to check code and proofread the manuscript. All content was reviewed and edited by the authors, who take full responsibility for the publication.

%% file: Appendix1_unmarked.tex
The generative process of the standard Reinforcement Learning (RL) model is summarised as follows:

\begin{enumerate}
  \item Draw RL exploration-exploitation sensitivity parameter
        $\beta \;\sim\; \mathrm{Lognormal}\bigl(\mu_{\beta},\, \sigma_{\beta}\bigr)$,\\
        ensuring that $\beta > 0$.
  \item Draw RL learning-rate parameter 
        $z_{\alpha} \;\sim\; \mathcal{N}\bigl(\mu_{z_{\alpha}},\, \sigma_{z_{\alpha}}\bigr)$,
        \\
        then apply a sigmoid transformation to ensure a rate in \([0,1]\):
        \\
        $\alpha \;=\; S\bigl(z_{\alpha}\bigr) 
                       \;=\; \frac{1}{1 \;+\; e^{-\,z_{\alpha}}}$
  \item \textbf{For each alternative} $i \in \{1,\ldots,I\}$:
    \begin{enumerate}
      \item Draw RL bias parameters $\gamma_{i} \;\sim\; \mathcal{N}\bigl(\mu_{\gamma_{i}},\, \sigma_{\gamma_{i}}\bigr)$\\
            (One alternative is set to zero for identification purposes.)
      \item Draw initial expectations
            $z_{Q_{i0}} \;\sim\; \mathcal{N}\bigl(\mu_{z_{Q_{i0}}},\, \sigma_{z_{Q_{i0}}}\bigr)$,
            \\
            then apply the following sigmoid transformation to ensure positive initial $Q$ values in \([a,b]\):
            $Q_{i0} = a+(b-a) \times S\bigl(z_{Q_{i0}}\bigr)$, where a and b are fixed by the experimental design and are not estimated.
    \end{enumerate}

  \item \textbf{For each individual} $n \in \{1,\ldots,N\}$:
    \begin{enumerate}
      \item \textbf{For each choice occasion} $t \in \{1,\ldots,T_{n}\}$:
        \begin{enumerate}
              \item Draw observed choice
                    $y_{nt} \;\sim\; \mathrm{MNL}\bigl(\beta,\,\{\gamma_{i},\,Q_{nit}\}_{i=1}^I\bigr)$,
          \item Compute the prediction error of the chosen alternative $y_{nt} = j$ (according to step 4.(a).i):\vspace{-0.8em}
                \[\delta_{njt} 
                  \;=\; r_{njt} \;-\; Q_{njt}\]
          \item Update the expectation of the chosen alternative $y_{nt} = j$ (according to step 4.(a).i):\vspace{-0.8em}
                \[Q_{nj(t+1)} \;=\; 
                Q_{njt}
                \;+\; \alpha\,\delta_{njt}\]
        \end{enumerate}
    \end{enumerate}
\end{enumerate}

%% file: Appendix2_unmarked.tex
This appendix presents a synthetic validation and parameter recovery exercise to assess whether the proposed LCRL model and its variational inference procedure can recover known data-generating parameters under controlled conditions. Parameter recovery is a simulation-based validation technique that involves generating synthetic datasets from the model with known `true' parameters, estimating the model on these generated datasets, and then comparing the recovered estimates to the original values. Successful recovery provides evidence that the model is identifiable, the inference procedure is correctly implemented, the cognitive assumptions hold when applied to real data, and the cognitive and behavioural interpretation of the parameter estimates are valid and not artefacts of estimation or model specification. This exercise should be interpreted as a validation of parameter recovery and implementation, rather than as a full assessment of variational posterior calibration relative to exact Bayesian inference.

In this exercise, we generate 100 synthetic datasets that replicate the structure and experimental design of the real dataset from Section \ref{section:dataset} as well as the adaptations introduced in Section \ref{section:results}. Each simulated dataset uses the same number of respondents (83), the same socio-economic and demographic characteristics per respondent, the same number of trials per respondent (20), the same experimental setup (10 driving simulator choices and 10 stated preference choices), and the same choice setting between a 5-minute reliable route and an unreliable route that can take between 2 and 7 minutes. Using a two-class LCRL model, we draw a new set of class-specific parameters per simulation: baseline bias parameters for the reliable option ($\gamma_{DS,Rel,k}$) from a normal distribution with mean 0 and standard deviation 1; an SP-shift bias ($\gamma_{SP,Rel,k}$) from the same distribution; exploration-exploitation sensitivities for DS and SP contexts ($\beta_{DS,k}$ and $\beta_{SP,k}$) from a uniform distribution between 0.1 and 2; learning rates ($\alpha_{k}$) from a uniform distribution between 0.05 and 0.95; initial expected time for the unreliable route ($Q_{Unr,0,k}$) from a uniform distribution between 2 and 7; and class-membership parameters ($\eta_k$) from a normal distribution with mean 0 and standard deviation 1. For each respondent, the two-class LCRL model uses the new set of parameters to generate a full sequence of choices and feedback. This results in 100 synthetic panel datasets that are produced by known `true' parameters but mirror the structure of the real data. The simulated datasets are then fitted using the same LCRL specification and inference settings as in the real application in Section \ref{section:results}
(e.g., variational inference, priors, and constraints). 

\begin{figure}[H]   
    \centering
    \includegraphics[origin=c, width = 1.0 \textwidth
   ]{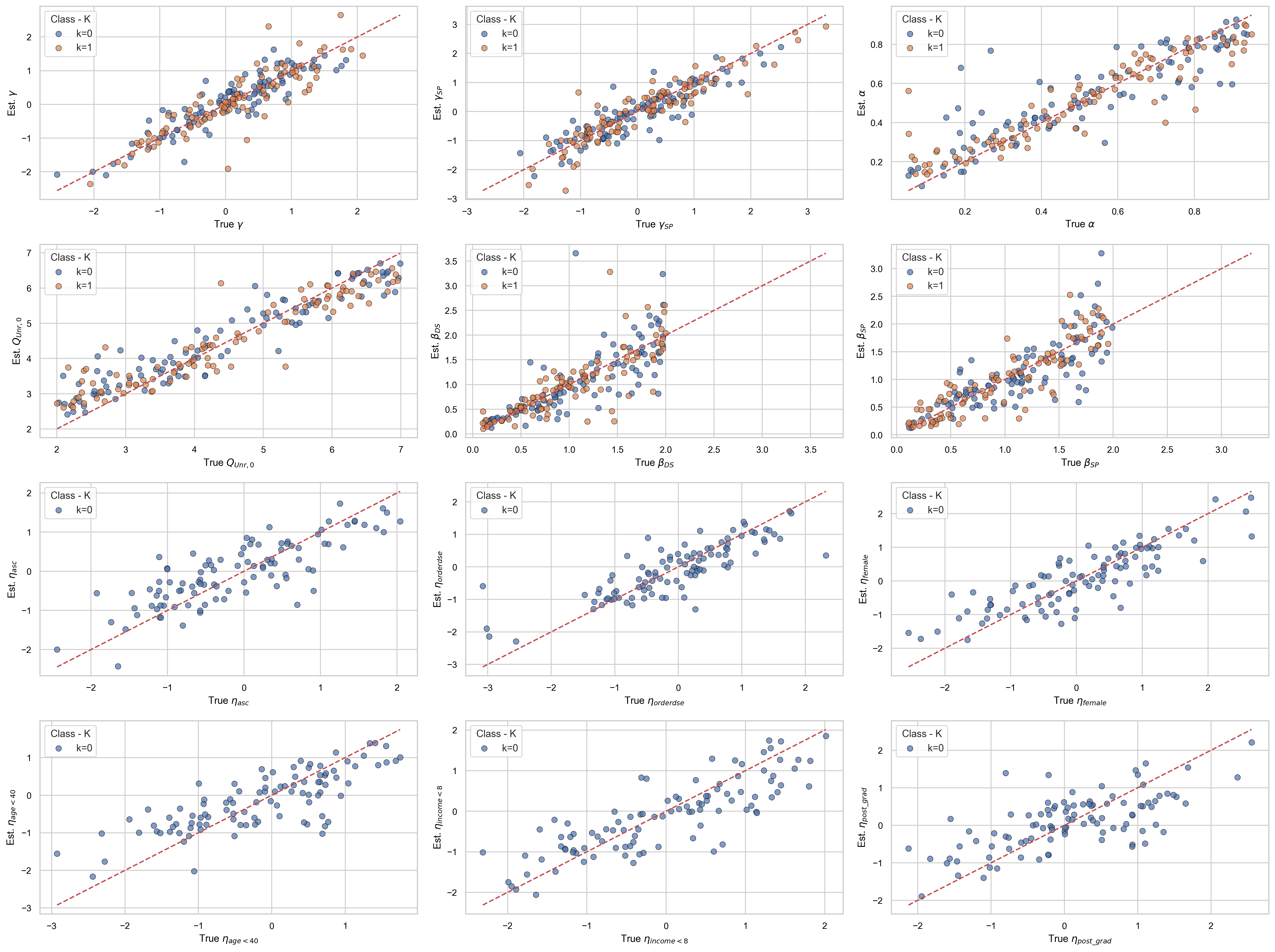}
    \caption{Parameter recovery plots of the LCRL model with two classes on simulated datasets}
    \label{fig:lcrl_recovery}
\end{figure}

Figure \ref{fig:lcrl_recovery} shows the posterior mean of each estimated parameter against its corresponding ground-truth value across the 100 simulated datasets. The plots indicate that the proposed LCRL model effectively recovers the underlying parameters that generated the data. The class-specific RL choice parameters are generally aligned around the 45-degree line, with only minor dispersion around it for most parameters. Although the learning rates are generally challenging to recover in short repeated choice sequences, the estimates closely align with the simulated values. Moreover, the context-specific bias parameters exhibit strong recovery and the sensitivity parameters a reasonable one, supporting the DS-SP separation in the model specification. Similarly, the initial expected times for the unreliable route show near-perfect recovery, indicating that the model successfully distinguishes between prior beliefs and subsequent updates.
The class-membership parameters, which link socio-economic and demographic attributes to latent classes, display slightly greater variation around the 45-degree line. This is expected, as their identifiability depends on the behavioural distance between classes in each simulation and some simulated parameter sets may produce classes that are relatively similar. Nevertheless, the plots indicate a reliable level of recovery overall. 

In addition to the qualitative visual inspection, we compute quantitative metrics that summarise how well each parameter is recovered across all simulations. For every parameter and latent class, we compute the average bias as the mean difference between the estimated and true values, the normalised root mean squared error (NRMSE), the correlation between the true and recovered values, and the coefficient of determination ($R^2$). All metrics are computed using the posterior means of the estimated parameters and their corresponding ground-truth values.

The NRMSE rescales the RMSE by the range of the true parameter values across the 100 simulated datasets, allowing comparisons across parameters with different scales:
\begin{equation}
    \text{NRMSE} = \frac{\sqrt{\frac{1}{S}\sum_{s=1}^{S}(\hat{\theta}_s - \theta_s)^2}}{\max(\theta_s) - \min(\theta_s)},
\end{equation}
where $S$ is the number of simulated datasets, $\theta_s$ are the ground-truth parameters, and $\hat{\theta}_s$ are the corresponding recovered posterior means. 

The coefficient of determination is defined as:
\begin{equation}
    R^2 = 1 - \frac{\sum_{s=1}^{S}(\hat{\theta}_s - \theta_s)^2}{\sum_{s=1}^{S}(\theta_s - \text{mean}(\theta_s))^2}
\end{equation}

The recovery statistics summarised in Table \ref{tab:param_recovery_lcrl} confirm that the proposed LCRL model can reliably recover the underlying data-generating parameters across the 100 simulated datasets. Bias values are close to zero for all parameters and the normalised RMSE values are generally low, indicating that estimation error is small relative to the range of true parameter variation. Correlations between the true and recovered estimates typically exceed 0.80, with $R^2$ values above 0.65 for most parameters, indicating that the variational inference procedure consistently reconstructs the structure of the original parameter space.

Notably, the class membership parameters show the lowest correlations, between 0.68 and 0.85, with \text{NRMSE} values in the 0.10--0.15 range and $R^2$ values between 0.45 and 0.72. However, as previously mentioned, their identifiability depends on the behavioural separation between classes in each simulation, and some simulation draws might produce similar or overlapping behavioural classes. Nevertheless, values still show a moderate but reliable recovery of such parameters.

Overall, the recovery results provide evidence that the proposed LCRL model can recover the main parameters that shape both learning dynamics and latent-class segmentation under the simulated designs considered. These findings support the credibility of the empirical analysis in Section \ref{section:results} while recognising that the exercise evaluates parameter recovery rather than exact posterior calibration.

\begin{figure}[H]   
    \centering
    \includegraphics[origin=c, width = 1.0 \textwidth
   ]{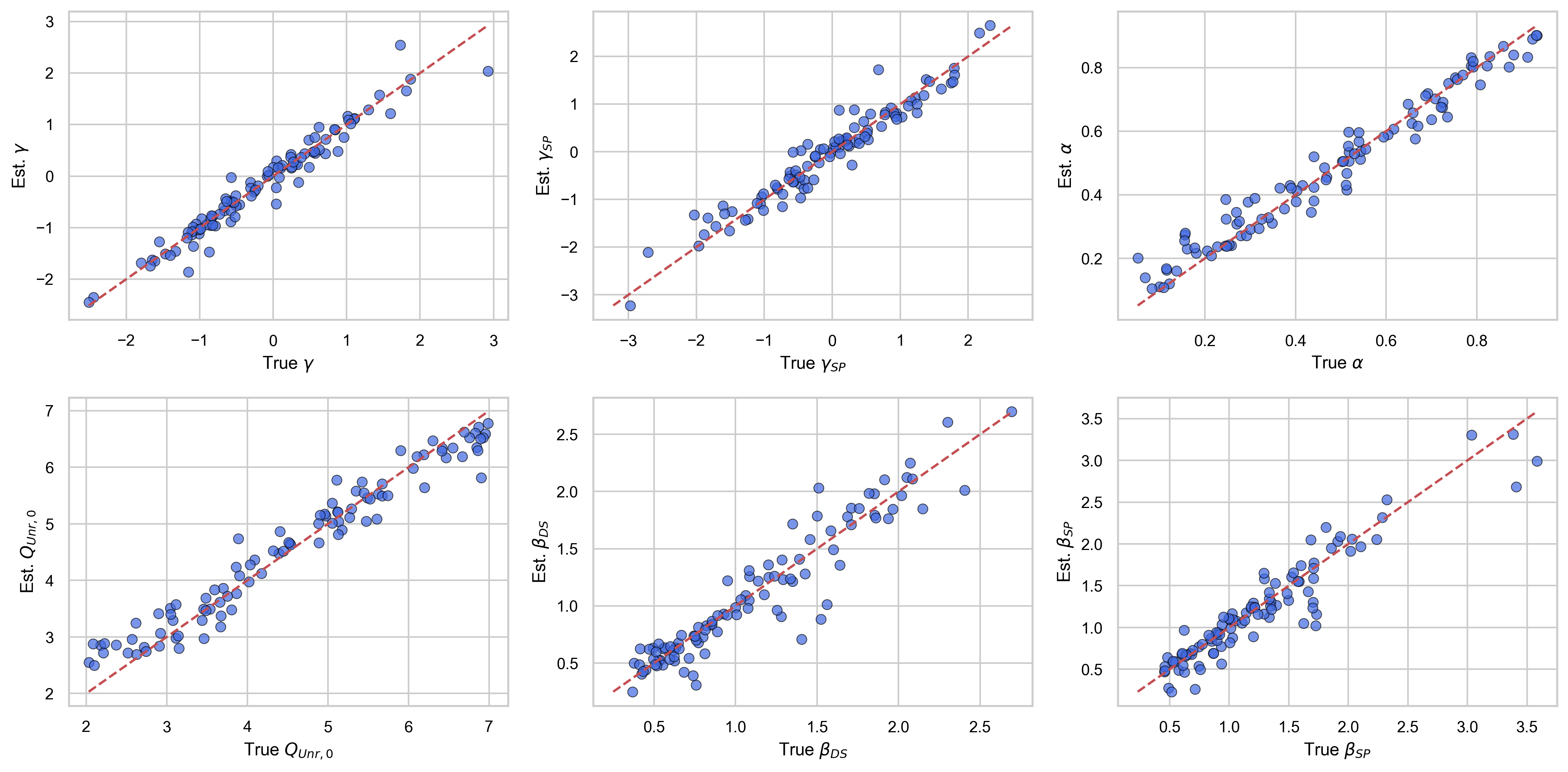}
    \caption{Parameter recovery plots of the RL model on simulated datasets}
    \label{fig:rl_recovery}
\end{figure}

\begin{table}[H]
\centering
\begin{threeparttable}
\caption{Parameter recovery metrics of the LCRL model on simulated datasets}
\label{tab:param_recovery_lcrl}
\begin{tabular}{l c r r r r r}
\toprule
\textbf{Parameter} & \textbf{k} & \textbf{Bias} & \textbf{NRMSE} & \textbf{Correlation} & \textbf{$R^2$} \\
\midrule
$\gamma_{\mathrm{DS,Rel,k}}$     & 0 & 0.018 & 0.086 & 0.901 & 0.804 \\
$\gamma_{\mathrm{DS,Rel,k}}$     & 1 & -0.038 & 0.107 & 0.883 & 0.746 \\
$\gamma_{\mathrm{SP,Rel,k}}$       & 0 & -0.062 & 0.096 & 0.893 & 0.790 \\
$\gamma_{\mathrm{SP,Rel,k}}$       & 1 & -0.060 & 0.083 & 0.916 & 0.828 \\
$\alpha_{\mathrm{k}}$              & 0 & 0.014 & 0.141 & 0.880 & 0.772 \\
$\alpha_{\mathrm{k}}$              & 1 & 0.002 & 0.117 & 0.919 & 0.837 \\
$Q_{\mathrm{Unr,0,k}}$             & 0 & 0.120 & 0.105 & 0.949 & 0.882 \\
$Q_{\mathrm{Unr,0,k}}$             & 1 & 0.031 & 0.108 & 0.941 & 0.870 \\
$\beta_{\mathrm{DS,k}}$            & 0 & -0.019 & 0.238 & 0.781 & 0.420 \\
$\beta_{\mathrm{DS,k}}$            & 1 & -0.019 & 0.193 & 0.819 & 0.565 \\
$\beta_{\mathrm{SP,k}}$            & 0 & -0.025 & 0.194 & 0.790 & 0.490 \\
$\beta_{\mathrm{SP,k}}$            & 1 & 0.003 & 0.155 & 0.886 & 0.742 \\
$\eta_{\mathrm{ASC}}$              & 0 & 0.073 & 0.125 & 0.815 & 0.657 \\
$\eta_{\mathrm{DS\,First}}$        & 0 & -0.021 & 0.105 & 0.823 & 0.677 \\
$\eta_{\mathrm{Female}}$           & 0 & -0.006 & 0.109 & 0.848 & 0.719 \\
$\eta_{\mathrm{Age<40}}$           & 0 & 0.014 & 0.135 & 0.781 & 0.609 \\
$\eta_{\mathrm{Income<80k}}$       & 0 & -0.092 & 0.137 & 0.832 & 0.685 \\
$\eta_{\mathrm{PostGrad}}$         & 0 & 0.016 & 0.150 & 0.678 & 0.453 \\
\bottomrule
\end{tabular}
\end{threeparttable}
\end{table}

\begin{table}[H]
\centering
\begin{threeparttable}
\caption{Parameter recovery metrics of the RL model on simulated datasets}
\label{tab:param_recovery_rl}
\begin{tabular}{l c r r r r}
\toprule
\textbf{Parameter} & \textbf{Bias} & \textbf{NRMSE} & \textbf{Correlation} & \textbf{$R^2$} \\
\midrule
$\gamma_{\mathrm{DS}}$                      & -0.024 & 0.041 & 0.973 & 0.946 \\
$\gamma_{\mathrm{SP}}$        & 0.023 & 0.052 & 0.965 & 0.930 \\
$\alpha$                      & 0.006 & 0.057 & 0.981 & 0.958 \\
$Q_{\mathrm{Unr,0}}$          & 0.032 & 0.068 & 0.977 & 0.946 \\
$\beta_{\mathrm{DS}}$         & -0.020 & 0.082 & 0.944 & 0.880 \\
$\beta_{\mathrm{SP}}$         & -0.042 & 0.069 & 0.944 & 0.886 \\
\bottomrule
\end{tabular}
\end{threeparttable}
\end{table}

In addition, we perform parameter recovery analysis for the benchmark RL model, presented in Section \ref{section:results}, using the same simulation applied for the LCRL model. Figure \ref{fig:rl_recovery} shows near-perfect recovery for all RL parameters, as all values are well clustered around the 45-degree line. Similarly, Table \ref{tab:param_recovery_rl} indicates exceptionally strong recovery across all parameters. Bias values are near zero, and NRMSE values remain very low (less than 0.1), reflecting minimal estimation error relative to the scale of the simulated parameters. Correlations exceed 0.94 for all parameters, with $R^2$ values above 0.88, indicating that the recovered estimates closely track the true data-generating values.